\documentclass[journal]{IEEEtran}
\usepackage{orcidlink}
\usepackage{amsmath}
\DeclareUnicodeCharacter{062C}{}
\DeclareUnicodeCharacter{0645}{}
\DeclareUnicodeCharacter{0629}{}
\DeclareUnicodeCharacter{0647}{}
\DeclareUnicodeCharacter{0646}{}
\DeclareUnicodeCharacter{0644}{}
\DeclareUnicodeCharacter{0627}{}
\DeclareUnicodeCharacter{0639}{}
\DeclareUnicodeCharacter{0631}{}

\begin{document}
 
\title{Energy-Efficient Fast Object Detection on Edge Devices for IoT Systems}

\author{\IEEEauthorblockN{Mas Nurul Achmadiah\orcidlink{0000-0002-2162-9575} , Afaroj Ahamad\orcidlink{0000-0002-6261-8821}, Chi-Chia Sun \orcidlink{0000-0002-3112-2516}~\IEEEmembership{Member,~IEEE,} Wen-Kai Kuo \orcidlink {0000-0002-7872-483X}~\IEEEmembership{Member,~IEEE,}
\IEEEauthorblockA{}
\thanks{Mas Nurul Achmadiah is with the SMIM Research Center and the Department of Electro-Optics, National Formosa University, Yunlin, Taiwan; Wen-Kai Kuo is with the Department of Electro-Optics, National Formosa University, Yunlin, Taiwan; Afaroj Ahamad is with Department of Computer Science and Engineering at Yuan Ze University, Taoyuan, Taiwan; Chi-Chia Sun is with the Department of Electrical Engineering, National Taipei University, Taipei, Taiwan;\\ Corresponding author is Chi-Chia Sun (\textit{E-mail: chichiasun@mail.ntpu.edu.tw})}}}

\markboth{IEEE INTERNET OF THINGS JOURNAL, Vol.xx, No.x, xxxxxx-xxxx}
{Achmadiah \MakeLowercase{\textit{et al.}}: An Improved Fast-Moving Object Detection Using Frame Difference With AI Classifier}

\maketitle
\begin{abstract}
This paper presents an Internet of Things (IoT) application that utilizes an AI classifier for fast-object detection using the frame difference method. This method, with its shorter duration, is the most efficient and suitable for fast-object detection in IoT systems, which require energy-efficient applications compared to end-to-end methods. We have implemented this technique on three edge devices: AMD Alveo$^{TM}$U50, Jetson Orin Nano, and Hailo-$8^{TM}$AI Accelerator, and four models with artificial neural networks and transformer models. We examined various classes, including birds, cars, trains, and airplanes. Using the frame difference method, the MobileNet model consistently has high accuracy, low latency, and is highly energy-efficient. YOLOX consistently shows the lowest accuracy, lowest latency, and lowest efficiency. The experimental results show that the proposed algorithm has improved the average accuracy gain by 28.314\%, the average efficiency gain by 3.6 times, and the average latency reduction by 39.305\% compared to the end-to-end method. Of all these classes, the faster objects are trains and airplanes. Experiments show that the accuracy percentage for trains and airplanes is lower than other categories. So, in tasks that require fast detection and accurate results, end-to-end methods can be a disaster because they cannot handle fast object detection. To improve computational efficiency, we designed our proposed method as a lightweight detection algorithm. It is well suited for applications in IoT systems, especially those that require fast-moving object detection and higher accuracy.
\end{abstract}

 
\IEEEpeerreviewmaketitle

\section{Introduction}
\label{sec:introduction}
\IEEEPARstart{T}he Internet of Things (IoT) is crucial for advancing computer vision by enabling seamless integration of sensors, devices, and data processing. The global proliferation of IoT over the past decade has facilitated the development of numerous new applications that utilize a wide range of devices and sensors. With the explosive growth of mobile applications and rapid development of advanced wireless technologies under the driving forces of IoT networks, a large number of IoT devices and emerging applications require ultra data rates, low-energy consumption, and spectral efficiency simultaneously \cite{RR2}. The importance of improving efficiency in IoT applications is to overcome limitations in traditional systems by optimizing resource utilization and energy consumption in a way that optimization can provide innovative solutions to improve the performance of IoT systems in challenging environments \cite{kk1}. More recently, visual sensors have seen their considerable boom in IoT systems because they are capable of providing richer and more versatile information. IoT networks connect cameras and sensors to edge devices, facilitating real-time data collection and analysis. This connectivity enhances the capabilities of computer vision systems, making them more efficient in tasks like object detection, facial recognition, and anomaly detection. The massive data generated by IoT devices can be processed and analyzed to train and improve machine learning models, driving innovation in automation, smart cities, healthcare, and security \cite{r1}.

The field of moving object detection is so vast. Researchers have proposed many approaches to the importance of efficiency and accuracy in multitask learning by highlighting that integrating tasks such as semantic segmentation and camera pose prediction within a single framework not only improves the learning efficiency but also enhances the generalization ability across tasks \cite{kk2}. A critical task in object detection is motion estimation. Optical flow is commonly used to estimate the object's motion \cite{reff7}. Starting with the original algorithms by Lucas and Kanade (LK) \cite{reff8} as well as Horn and Schunck (HS) \cite{reff9}, gradient-based methods have led to other improved optical flow estimation methods. However, when the image background is cluttered or the detected object is moving at high speed, the accuracy of gradient-based methods will be significantly decreased.

There are three primary categories of techniques for detecting moving objects. One technique is optical flow \cite{reff4}, which establishes an image's optical flow field and looks at the associated pixel's motion vector to identify moving objects. On the other hand, because of the scene's variability, the calculation is intricate and prone to detection errors \cite{reff6}. However, the end-to-end method offers powerful capabilities and has achieved remarkable success, but it comes with several significant disadvantages. These include high computational and data requirements, lack of interpretability, risks of overfitting, scalability challenges, sensitivity to architecture and hyperparameters, ethical concerns, and maintenance difficulties. Addressing these issues requires advanced techniques, careful design, and ongoing management. Therefore, it cannot deliver optimal performance at high resolutions \cite{ref21}.

The following method is called frame difference \cite{reff12}. Detection of moving objects from a sequence of frames captured from a static camera is widely performed by the frame difference method. The objective of the approach is to detect the moving objects from the difference between the existing and reference frames \cite{reff13}. The frame difference method is the standard method of motion detection. This method adopts pixel-based differences to find the moving object. Which compares the pixel information among adjacent frames. When an object passes through the frames, the differences between the frames exceed the threshold \cite{kk4}. From \cite{reff14} the study results show the success of the frame difference method, which is more accurate in detecting objects. In the frame difference method, the high resolution of the resulting video does not reduce the resulting performance.

Another essential task in object detection is classification. Real-time classification of fast-moving objects is a challenging task. Although image-based object classification systems have been explored for decades, classifying fast-moving objects in real-time and for a long duration is still challenging. Fast-moving objects often generate dramatic motion blurs in the images captured. The induced motion blurs cause severe image quality degradation, reducing the achievable classification accuracy. In addition, sophisticated image analysis algorithms are generally computationally expensive, and consequently, they are not suited for real-time classification \cite{ref17}.

Many industrial, medical, commercial, and research-related
applications depend on computer vision and image processing
techniques for real-time object recognition and classification. Central Processing Units (CPUs) are insufficient for many applications because they cannot quickly process the computations. Algorithms can be implemented in AI accelerators, Field-Programmable Gate Arrays (FPGAs), or Graphics Processing Units (GPUs) to shorten the calculation time. Choosing the right hardware accelerator for a given application can be difficult \cite{reff18}.

Several generations of FPGAs, GPUs, and AI accelerators
are available, and it is challenging to compare different hardware accelerators due to their technological variations. Previous works have covered the performance and technical aspects of hardware accelerators. Nevertheless, many of these presentations have flaws, such as discussing outdated technology and comparing hardware accelerators at two distinct technological levels without providing enough technical information to help choose an appropriate accelerator \cite{reff20}. Currently, the most commonly used hardware is the GPU. However, GPUs have significant power consumption and high latency. Some AI processing hardware with low latency, high throughput, and low power include AI Accelerator, FPGA, and Special AI SoC.

Energy efficiency is crucial for edge devices in the IoT due to their typically limited power resources and the vast scale at which they operate \cite{RR3}. Edge devices, such as sensors, gateways, and microcontrollers, are often deployed in remote or hard-to-reach locations where frequent battery replacement or recharging is impractical. Therefore, optimizing energy usage is essential to prolong device lifespan, reduce maintenance costs, and ensure the reliability of the entire IoT ecosystem. Efficient energy management in edge devices also contributes to the overall sustainability of IoT networks, reducing their carbon footprint and supporting green technology initiatives. Techniques such as low-power hardware design and intelligent power management algorithms are critical for minimizing energy consumption \cite{r22}.

The main contribution of this paper is to compare our proposed method to the end-to-end method and implement this technique on three modern edge devices for object detection and classification applications to get the best method.
Specifically, we make the following contributions: 
\begin{enumerate}
\item Comparing our proposed object detection method with end-to-end methods and finding which method has real-time processing, high accuracy, and is highly energy-efficient.
\item Perform testing on different edge devices; first, we tested on the AMD Alveo U50. Alveo is an accelerator product from AMD that uses FPGA technology. The second is Jetson Orin Nano, and the third is Hailo-$8^{TM}$ AI Accelerator. 
\item From these two results, we will compare the performance of the method and the capabilities of edge devices in the context of object detection. We categorize and report our evaluation results based on key benchmarks, including object detection accuracy, latency performance, and efficiency gains.
\end{enumerate}
This paper is organized as follows. Section 1 discusses the background of the proposed method. Part 2 discusses related work, which includes previous research. Section 3 discusses the proposed method and the steps required; Section 4 shows the experimental results of the evaluated devices, which are then analyzed according to the evaluation criteria. Finally, Section 5 summarizes our work and concludes.

\section{RELATED WORK}
In this section, we survey the previous literature related to our research. This section involves two parts. The first discusses our proposed algorithm, Fast-Moving Object Detection (FMOD), and AI Classifier. The second concerns the implementation of FMOD on the AMD Alveo U50, Jetson Orin Nano, and Hailo- $8^{TM}$ AI accelerators.

\subsection{Fast-Moving Object Detection (FMOD) and AI Classifier}
Real-time object detection is crucial for latency-sensitive IoT applications, such as autonomous driving, augmented reality, and intelligent surveillance. These applications require fast and accurate processing of video streams to make timely decisions. Traditional object detection methods, while accurate, often have high end-to-end latency, making them unsuitable for real-time use \cite{new2}. Object detection and the capacity to cross current benchmarks remain unresolved issues in deep learning technology. Gradually raising CNN's computational capacity promotes extensive application  \cite{11} . The CNN-based  pedestrian and localization algorithm proposed that the object was spotted using a monovision camera, and the distance of the found object was measured. Applied to classify the found objects \cite{12}. In \cite{13} presented a LiDAR-based three-stage GC-net encompassing a pipeline comprising grinding, clustering, and a CNN-based classifier. Leveraging the intelligence trend, Cooperative Driving Automation (CDA) has attended acknowledged events over the last several years. 

The first real-time object detection technique was proposed by \cite{21}. However, not all past methods are suitable for fast-moving object recognition and identification. Several investigations have successfully and highly precisely identified fast objects with great speed \cite{22}. The acoustic approach, however, failed to recover absolute visual and relative position. Regarding the image processing technique, fixed cameras are primarily applied in many types of research for fast object identification moving  \cite{23}. Though low accuracy and processing speed are significant disadvantages, this study effectively identified and classified objects at high speeds. Other earlier research found moving objects among more intricate backdrops with success. They still need to categorize the objects \cite{24}.

The frame differencing method is a straightforward and energy-efficient approach to motion detection that compares pixel intensity differences between consecutive video frames. By examining changes in pixel values, frame differencing directly identifies motion, avoiding the need for extensive preprocessing or feature extraction. Pixels with significant changes are flagged as areas of motion. This minimalistic processing pipeline avoids the computational overhead associated with more complex methods, such as convolutional neural networks (CNNs) used in YOLO or other deep learning models. In \cite{nov1} frame differencing can efficiently handle simple scenes with static backgrounds. While advanced models require extensive hardware and energy resources to train and infer object categories, frame differencing’s task-specific design achieves its objectives with significantly lower energy demands. Compared to background subtraction methods, frame differencing excels in terms of energy efficiency because it does not rely on maintaining or updating a background model. Background subtraction methods often require dynamic adaptation to changing environments, which introduces additional computational steps. In contrast, frame differencing processes only two consecutive frames at a time, making it lightweight and adaptable. This advantage is highlighted in the work of \cite{nov2} where the authors demonstrate the method’s ability to achieve real-time motion detection with minimal hardware requirements.

When compared to deep learning-based methods, the energy efficiency of frame differencing becomes even more apparent. Deep learning models like YOLO require extensive computations across multiple convolutional layers, resulting in high power consumption. These methods also demand specialized hardware such as GPUs or TPUs to operate in real-time, further increasing energy costs. In contrast, frame differencing can run on low-power microcontrollers without the need for hardware accelerators. In \cite{nov3} underscore this advantage, emphasizing frame differencing’s suitability for resource-constrained environments.

\subsection{Implementation of FMOD on edge devices for energy efficiency in IoT systems}
Fast-moving object detection, particularly object detection, is critical for applications requiring real-time analysis and immediate response. Moving object detection enables systems to identify and track objects in dynamic environments quickly, which is essential for applications like autonomous driving, where vehicles must rapidly detect and react to other cars, pedestrians, and obstacles to avoid accidents \cite{n1}. In industrial automation, fast object detection allows robots to monitor and interact with swiftly moving items on production lines, enhancing efficiency and precision. Security systems benefit from fast detection by immediately identifying and responding to potential threats, ensuring timely interventions \cite{n2}. In \cite{new1}, the development of energy-efficient Artificial Intelligence (AIoT) with intelligent edge computing is discussed. This paper highlights the importance of optimizing energy consumption in edge devices and cloud services when processing AIoT tasks. This paper introduces a multilevel intelligent edge framework designed to improve energy efficiency by managing resources between edge devices and the cloud.

Edge devices are indispensable in optimizing the performance and practicality of fast object detection. By processing data locally on devices such as smart cameras, drones, and mobile phones, edge computing significantly reduces latency, enabling immediate decision-making crucial for real-time applications \cite{n3}. This local processing minimizes the need for data transfer to remote servers, conserving bandwidth and ensuring that the system remains operational even with intermittent connectivity. Edge devices also enhance data privacy by keeping sensitive information on the device. Their role in fast object detection is vital in scenarios where speed and reliability are paramount, making AI applications more efficient, responsive, and capable of functioning independently of constant cloud connectivity \cite{n4}.

CPUs and GPUs each have strengths and weaknesses, influencing their suitability for various computing tasks. CPUs are versatile and excel in single-threaded performance, making them ideal for general-purpose computing and tasks that require complex decision-making and low latency. However, they need help with highly parallel computations due to their limited number of cores and higher power consumption. In contrast, GPUs are designed for parallel processing with thousands of smaller cores, making them excellent for tasks like image processing, scientific simulations, and deep learning. Despite their high throughput and efficiency in handling large-scale parallel tasks, GPUs consume significant power, have higher latency for single-threaded tasks, and are often expensive. However, GPU platforms are power hungry, and due to their high cost \cite{26}, ASIC's time-to-market weakness is unacceptable on AI Accelerator edge computing devices \cite{27} .

Hailo AI accelerators offer a specialized solution to address the limitations of both CPUs and GPUs, especially in edge AI applications. These accelerators are optimized for deep learning inference, providing high performance and power efficiency in a compact form factor. They are suitable for real-time AI tasks in low-power environments like IoT devices and autonomous systems. While Hailo AI accelerators offer significant advantages in power efficiency and latency for AI inference, they specialize in these tasks and may require additional integration effort and optimization knowledge. Overall, Hailo AI accelerators complement the capabilities of CPUs and GPUs by providing an efficient and effective solution for specific AI-driven applications \cite{link2}. FPGAs are highly configurable, allowing for custom hardware configurations optimized for specific tasks, leading to significant performance and power efficiency improvements. FPGAs can be tailored to handle parallel processing tasks with lower latency and power consumption than GPUs, making them suitable for applications where these factors are critical, such as real-time signal processing, embedded systems, and specialized AI inference tasks. Additionally, FPGAs can be reprogrammed to adapt to evolving computational requirements, providing a level of flexibility that fixed-function ASICs (Application-Specific Integrated Circuits) lack \cite{31}.

However, the complexity of FPGA programming and integration can be a disadvantage, requiring specialized knowledge and development time to leverage their capabilities thoroughly. Despite this, the adaptability and efficiency of FPGAs make them an essential component in scenarios where traditional CPUs and GPUs fall short. Apart from this, FPGA \cite{28}  has properties such as low operation power, reconﬁgurability, customizable data ﬂow, and data width. Hence, FPGA became a delightful platform for accelerating DNN architecture. In \cite{29} suggested an FPGA scalable processor, i.e., a deep learning accelerator unit (DLAU), in which they use a technique such as FIFO buffer and pipelines. This approach offers intercommunication reduction to DRAM while enabling the reuse of computing units when implementing neural networks. In \cite{30} suggested a large-scale CNN implementation methodology that involved constructing manifold tiles inside an FPGA processor. However, due to the limitation of FPGA hardware, most prior studies focus on simplifying the neural network weight coefﬁcient or architecture \cite{32}.  
 
It quantizes the parameters of the neural network to reduce the capacity required by memory and the logic gates needed by the architecture. Several studies have proposed methods to reduce memory and logic resource requirements with a slight reduction in accuracy in exchange \cite{33}. There are also prior studies enhancing the power consumption efficiency and performance of FPGA parallel processing with neural network architecture \cite{35}. However, most prior research did not discuss and consider practical applications. Moreover, it is usually 
challenging to select and detect objects within the input frame directly; therefore, this research paper will also propose an algorithm to detect objects of interest within the input frame.

Furthermore, the NVIDIA Jetson Orin Nano integrates a powerful CPU and GPU architecture tailored for AI workloads, providing efficient parallel processing and real-time performance with lower power consumption than traditional GPUs. This makes it particularly well suited for AI inference tasks at the edge, where power efficiency and compact form factor are critical. Additionally, the Jetson Orin Nano includes AI-specific accelerators and supports a wide range of AI frameworks, simplifying the deployment of advanced AI models in edge devices. By leveraging the capabilities of the Jetson Orin Nano, developers can achieve high-performance AI processing with improved energy efficiency and reduced latency, overcoming the traditional limitations of CPUs and GPUs. This makes it an excellent choice for applications such as autonomous machines, robotics, intelligent cameras, and other AI-driven embedded systems where computational power and energy efficiency are paramount \cite{youvan2024}.

The NVIDIA Jetson Orin Nano offers significant advantages over Hailo AI accelerators and FPGAs, particularly for edge AI applications. Unlike Hailo AI accelerators, the Jetson Orin Nano benefits from NVIDIA's extensive software ecosystem, including the JetPack SDK and support for popular AI frameworks such as TensorFlow and PyTorch. This makes development more straightforward and accessible. It combines CPU, GPU, and AI accelerators on a single chip, providing a versatile and integrated solution that can handle various tasks, from AI inference to general-purpose computing \cite{youvan2024}. Compared to FPGAs, the Jetson Orin Nano is more accessible to program and deploy, thanks to NVIDIA's high-level programming languages and comprehensive development tools. It is optimized specifically for AI workloads, offering efficient performance for neural network inference and real-time processing. Additionally, its power efficiency and ability to handle multiple tasks simultaneously make it ideal for edge applications with critical performance and power constraints. The Jetson Orin Nano's blend of powerful hardware, extensive software support, and ease of use makes it a compelling choice for implementing advanced AI capabilities at the edge. This research will apply the proposed method to each device so that the results can determine the best combination of evaluations, including object detection accuracy, latency performance, and increased efficiency \cite{Nv2}.

\section{THE PURPOSED METHOD}
The frame difference method combined with a lightweight AI classification algorithm is well-suited for fast-moving object detection in energy-constrained IoT applications due to its efficiency and performance. This method is lightweight, making it ideal for edge devices with limited power availability, such as sensors, cameras, and IoT gateways. By combining an efficient AI classification algorithm, this approach improves detection accuracy while maintaining low latency and energy consumption.

\begin{figure}[!ht]
\centering
\includegraphics[width=0.55\textwidth]{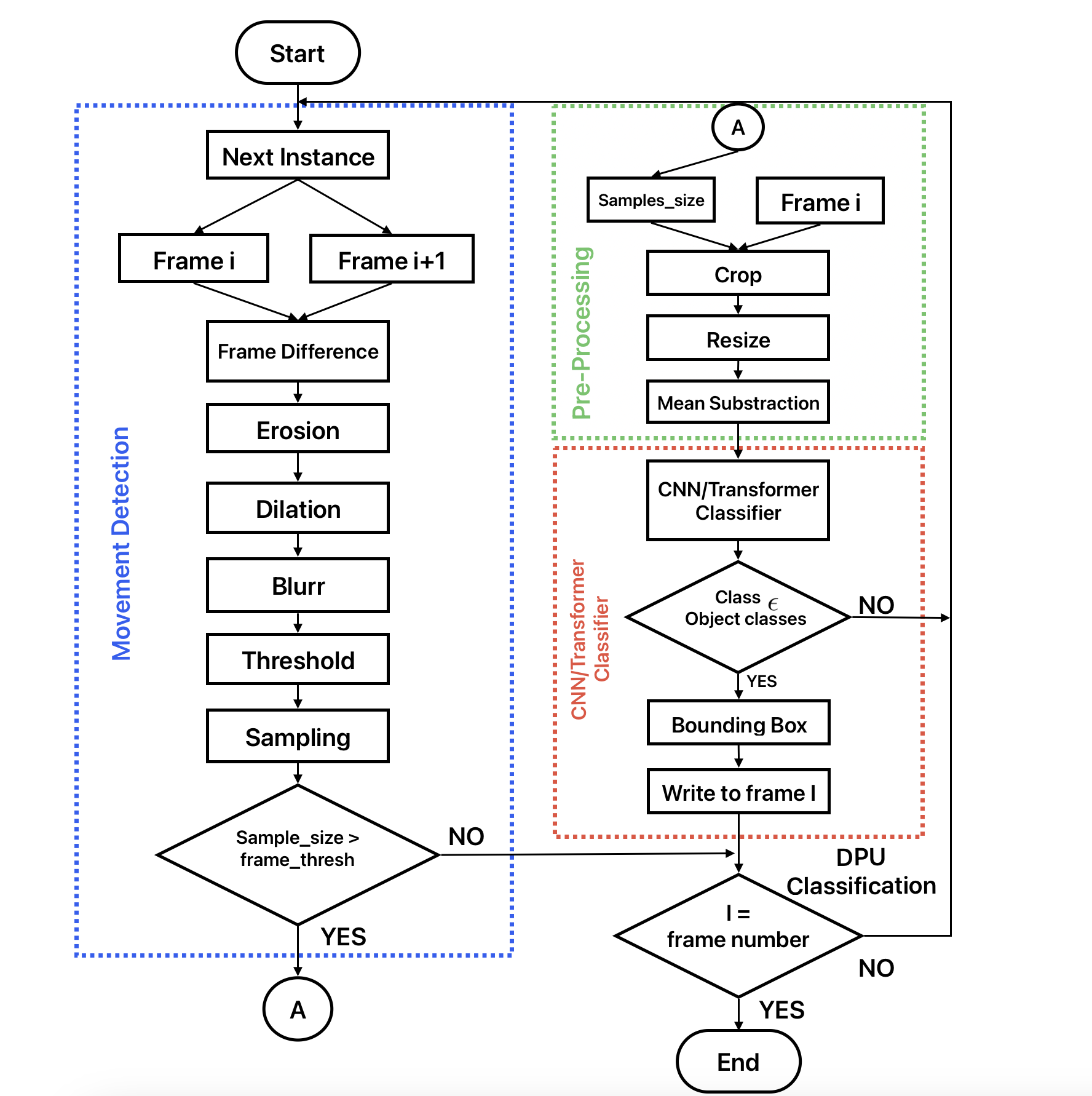}
\caption{Flow-Chart of Proposed Algorithm}
\label{fig:FC}
\end{figure}

This research involved three significant procedures: Movement Detection, Pre-processing, and CNN or Transformer Classifier. Fig.\ref{fig:FC} illustrated the proposed algorithm. Although Movement Detection and Pre-processing procedures can be grouped, as they both employ and are based on Image Morphology techniques, dividing them into two methods is preferable because a decision check separates them, contributing excellent features to the algorithm flow. The algorithm has a loop-like structure, presenting the complete loop of processing the 3-channel color video input and output of the resulting video, including features. The image region in which the object’s presence is marked with a bound box, and the class of the object is written above the box. Each video frame input represents a loop instance element of the algorithm. The algorithm starts and runs through a loop instance. Afterward, at the end of the loop instance, we will check if the current frame counter has already reached the maximum value or not, which is the total number of frames of the inputted video. If the frame counter has not reached the absolute frame number, the algorithm returns to the starting point and processes the next loop instance.

\subsection{Movement Detection Method}

The movement detection process uses image morphology as its foundation. Image morphology is the method of extracting useful information from an image or a video frame using a step-by-step procedure.

\begin{table}[ht]
\begin{tabular}{l}
\hline
\begin{tabular}[c]{@{}l@{}}\textbf{Algorithm 1} Feed Image\end{tabular} \\ \hline
1: \textit{video} \textbf{read}  \textit{vFrame}  \hspace{1.2cm} $\to$ \textbf{Read a store a video frame} \\
2: \textit{Im1} $\gets $  \textit{vFrame} \\
3: \textit{Im2} $\gets $  \textit{vFrame} \\
4: \textbf{for} $f\gets $  0, \textit{frame} \textit{Number} \textbf{do} $\to$ Loop whole video frame \\
5: \hspace{0.5cm} \textit{Im1} $\gets $   \textit{Im2} \hspace{1.6cm} $\to$ \textit{First input image} \\
6:\hspace{0.5cm} \textit{video} \textit{read} \textit{vFrame} \hspace{0.8cm} $\to$ \textit{Read next video frame} \\
7:\hspace{0.5cm} \textit{Im1} $\gets $ \textit{vFrame} \hspace{1.18cm} $\to$ \textit{Second input image} \\
8:\hspace{0.5cm} $ImgProc(Im1,Im2)  \hspace{0.53cm} \to$  Execute image morphology \\
9: \textbf{end for}
\\ \hline
\end{tabular}
\end{table}

Algorithm 1 illustrated each image's morphological steps. It involves covertly 3-color-channeling two consecutive frames to grayscale and feeding them as input. 

\begin{figure}[h!]
\centering
\includegraphics[width=0.45\textwidth]{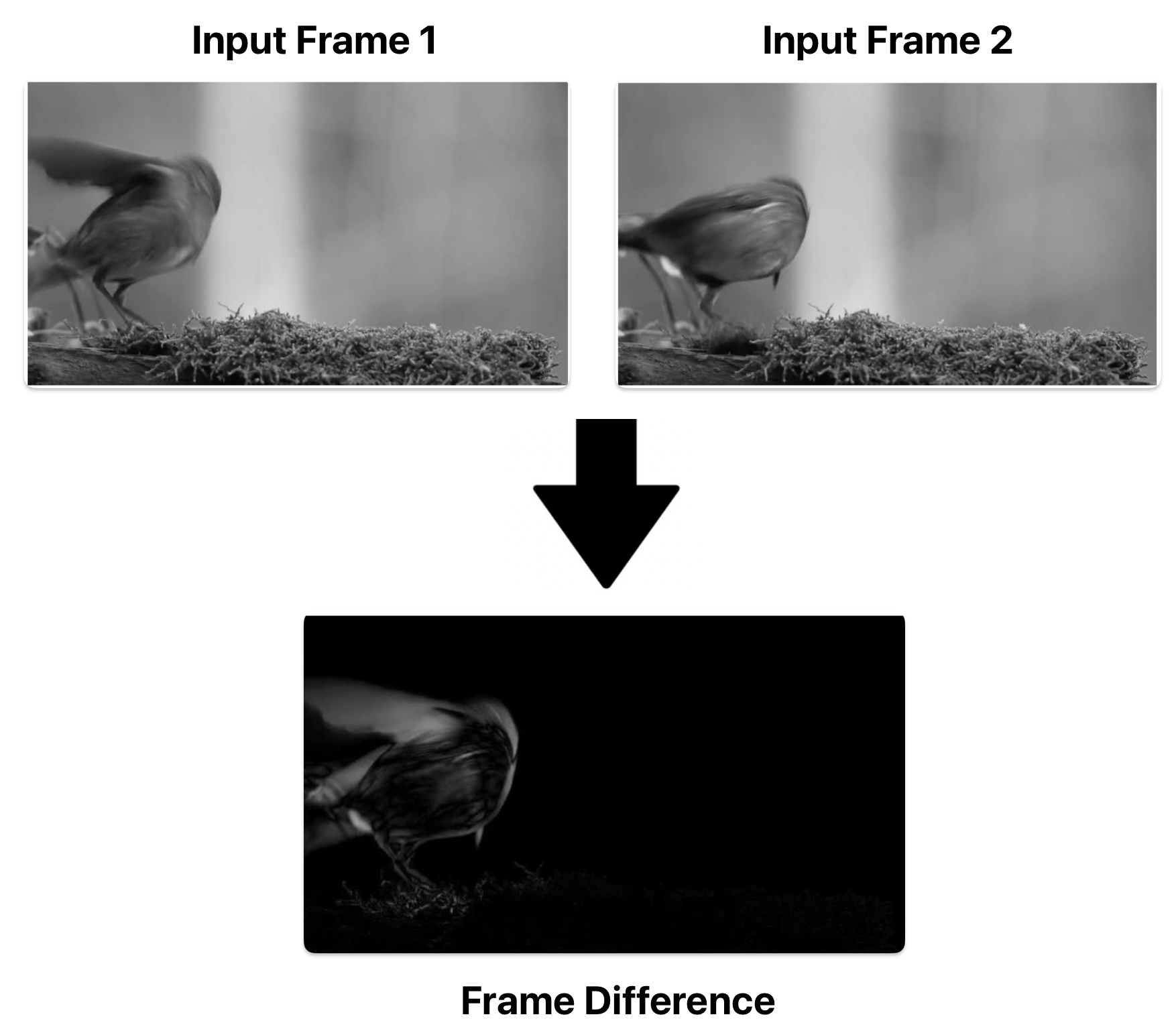}
\caption{Frame Difference Process}
\label{Fig:Framedif}
\end{figure}

The next step takes the frame's absolute abstraction of the 2 inputted images, i.e., frame difference, and gets 1 resulting image (see Fig. \ref{Fig:Framedif}). After the frame difference process, the image might already have a glimpse of which area contains the movement. However, the background still contributes a significant amount of noise. This noise may be caused by the subtle movement of the grass in the background, rather than by the object, and should thus be eliminated. For the goal of noise reduction, morphological opening is used. 

\begin{figure}[h!]
\centering
\includegraphics[width=0.41\textwidth]{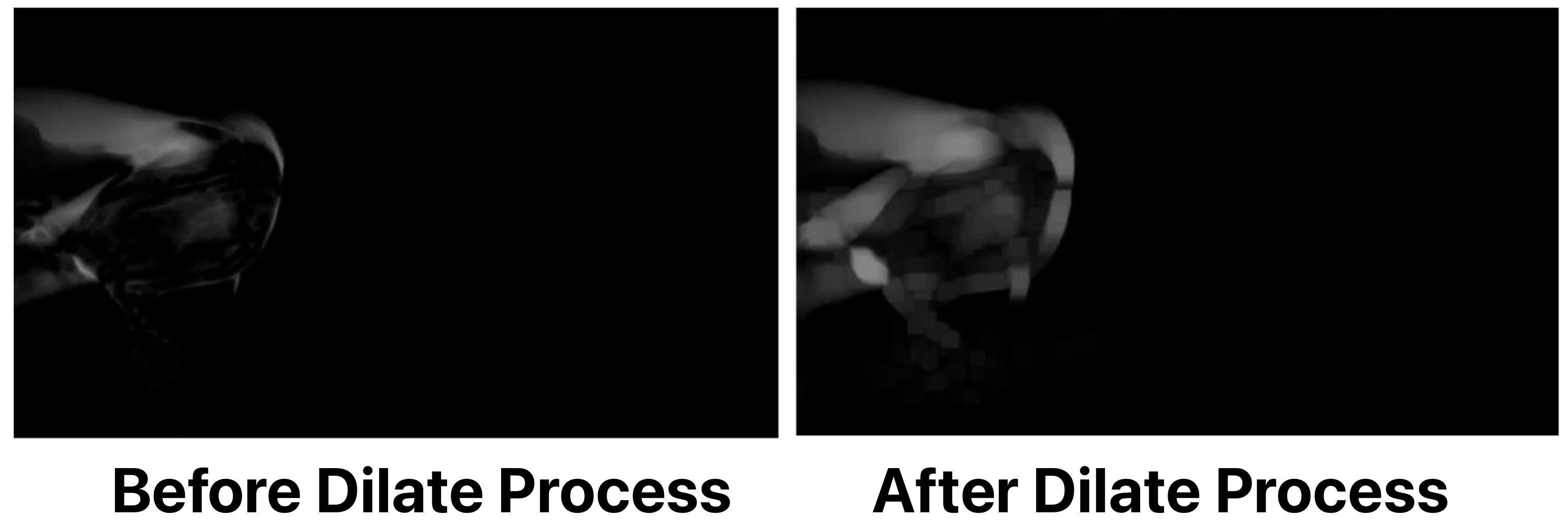}
\caption{Image Dilation Process}
\label{Fig:dilate}
\end{figure}

\begin{figure}[h!]
\centering
\includegraphics[width=0.41\textwidth]{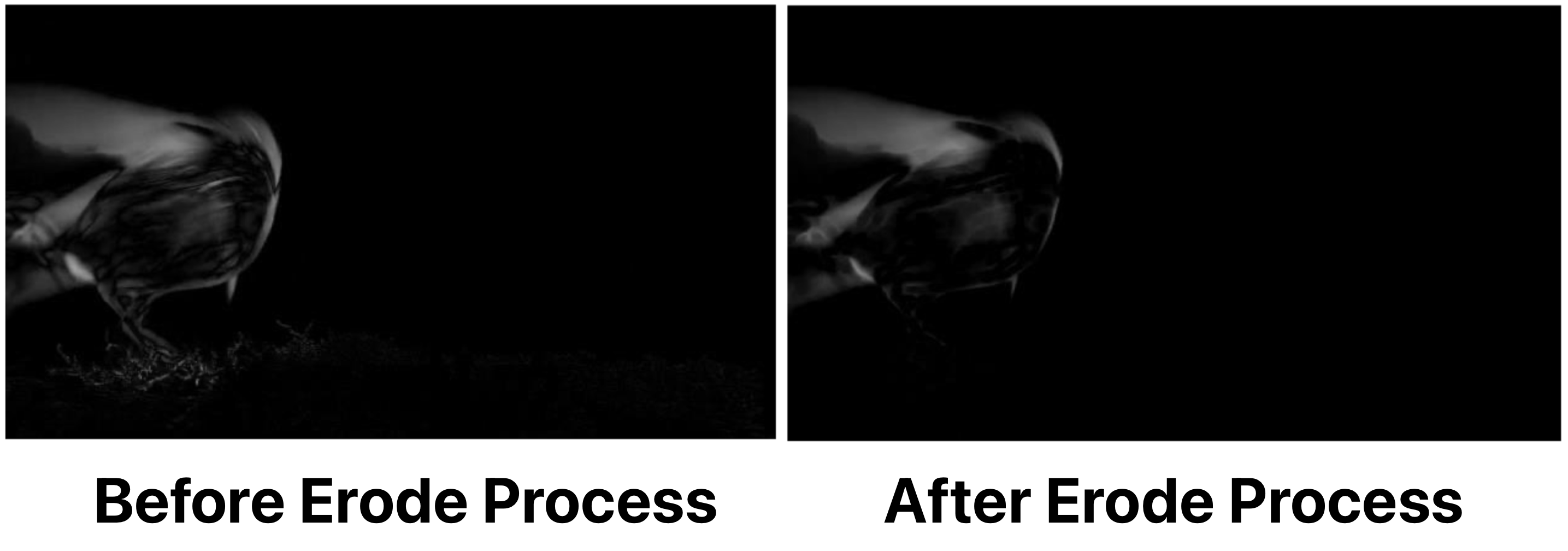}
\caption{Image Erosion Process}
\label{Fig:erode}
\end{figure}

The morphological opening of an image consists of 2 consecutive operations: an image erosion followed by an image dilation, both of which use the same structuring element. Fig. \ref{Fig:dilate} and \ref{Fig:erode} are the image dilation and erosion processes.

\begin{figure}[!ht]
\centering
\includegraphics[width=0.42\textwidth]{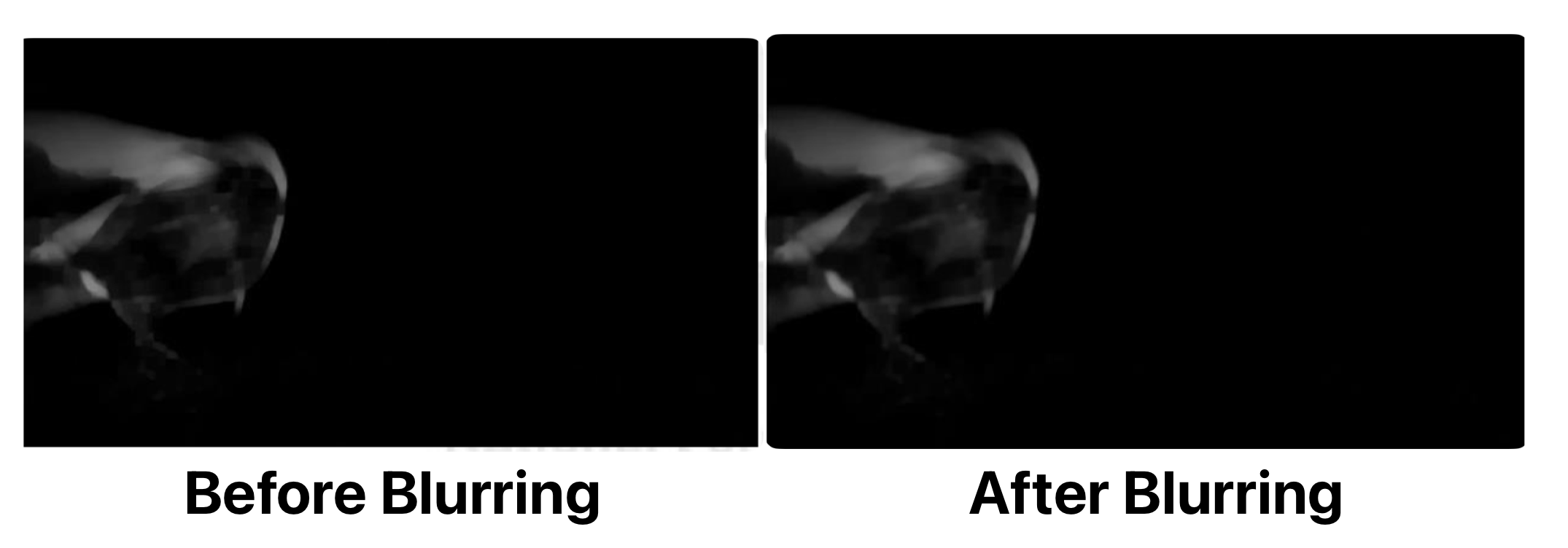}
\caption{Image Blurring Process}
\label{Fig:blur}
\end{figure}

\begin{figure}[!h]
\centering
\includegraphics[width=0.42\textwidth]{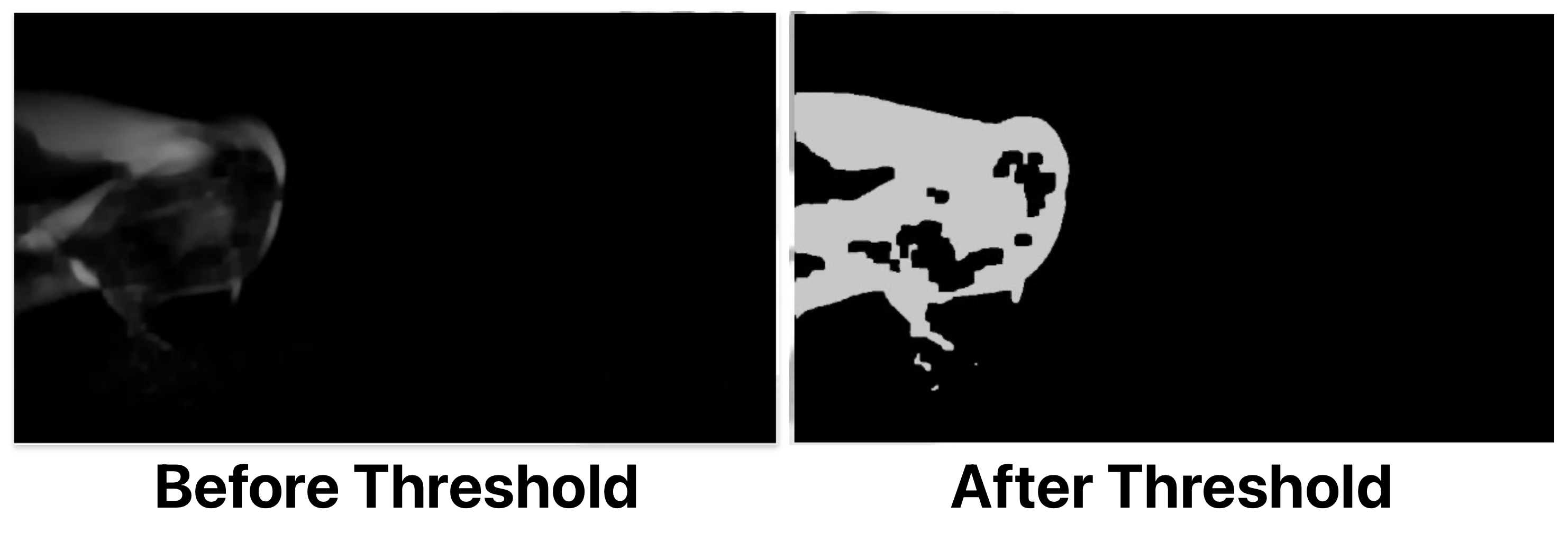}
\caption{Thresholding Process}
\label{Fig:thresh}
\end{figure}

The next step is called image blurring (see fig.\ref{Fig:blur}) the resulting image of the previous step is convoluted with a low-pass filter kernel of high-frequency content. After image blurring, the result is compared with the threshold value, and it is 0 for smaller than the threshold; otherwise, it is 255, i.e., it is called thresholding (see Fig. \ref{Fig:thresh}). The final step of image morphology is sampling, which involves finding the Region of Interest (ROI) rectangle from a frame.

Algorithm 2 illustrates pseudo-code to ROI. Firstly, find all points whose value equals 1, record their X and Y coordinates into the arrays, and determine the maximum and minimum of each array, respectively. Then the loop executes through both coordinate arrays one by one to feed maximum and minimum values, respectively. The algorithm will calculate the ROI's cropping parameters at the end.

\begin{table}[h!]
\begin{tabular}{l}
\hline
\textbf{Algorithm 2} Find ROI Arguments  \\ \hline
1: $X_{min}$ $\gets $ $maxRow$ - 1 $ \hspace{1cm} \to$  Initiate with opposite value         \\
2: $X_{max}$  $\gets $ 0                                               \\
3: $Y_{min}$  $\gets $ $maxCol$ - 1                                      \\
4: $Y_{max}$  $\gets $ 0                                               \\
5: \textbf{for} r  $\gets $ 0, $maxRow$ \textbf{do} \hspace{0.8cm} $ \to$ Search pixels with positive value \\
6: \hspace{0.5cm} \textbf{for} c $\gets $ 0, $maxCol$ \textbf{do}                                  \\
7: \hspace{1cm} \textbf{if} $Mat(r,c)$ = $True$  \textbf{then}                              \\
8:  \hspace{1.5cm} $X_{array}$ \textbf{append} c                                     \\
9: \hspace{1.5cm} $Y_{array}$ \textbf{append} r                                     \\
10: \hspace{1.5cm} $pointCnt$ $\gets $ $pointCnt$ + 1                               \\
11: \hspace{1cm} \textbf{end if}                                              \\
12: \hspace{0.5cm} \textbf{end for}                                             \\
13: \textbf{end for}                                          \\
14: \textbf{if} $pointCnt$ = 0 \textbf{then}                                \\
15: \hspace{0.5cm} $X_{position}$  $\gets $ 0                                        \\
16: \hspace{0.5cm} $Y_{position}$  $\gets $ 0                                        \\
17: \hspace{0.5cm} $X_{size}$  $\gets $ 0                                            \\
18: \hspace{0.5cm} $Y_{size}$  $\gets $ 0                                            \\
19: \textbf{return}        \hspace{2.7cm} $\to$  No point found                               \\
20: \textbf{end if}                                              \\
21: $X_{min} \gets X_{array}{[}0{]}  \hspace{1cm} \to$ Assign with first non-zero       \\
22: $X_{max}\gets X_{array}{[}0{]} $                                 \\
23: $Y_{min}\gets Y_{array}{[}0{]} $                                 \\
24: $Y_{max}\gets Y_{array}{[}0{]} $ \\
25: \textbf{for} $i$ $\gets 0, pointCnt \textbf{do} \hspace{0.5cm} \to$ Colomn searching loop \\
26: \hspace{0.5cm} \textbf{if} $X_{array}$ ${[}i{]}$  \textgreater $X_{max}$ \hspace{0.5cm} \textbf{then} \\
27: \hspace{1cm} $X_{min}$ $\gets$ $X_{array}$ ${[}i{]}$ \\
28: \hspace{0.5cm} \textbf{else}                                               \\
29: \hspace{1cm} \textbf{if} $X_{array}$ ${[}i{]}$ \textgreater $X_{max}$ \hspace{0.5cm} \textbf{then}                        \\
30: \hspace{1.5cm} $X_{max}$ $\gets$ $X_{array}$ ${[}i{]}$ \\
31: \hspace{1cm} \textbf{end if}                                             \\
32: \hspace{0.5cm} \textbf{end if}                                              \\
33: \textbf{end for}                                             \\
34: \textbf{for} i $\gets $ 0, $pointCnt$ \textbf{do} \hspace{0.5cm} $\to$  Row searching loop             \\
35: \hspace{0.5cm} \textbf{if} $Y_{array}$ ${[}i{]}$ \textless $Y_{min}$ \hspace{0.5cm} \textbf{then}         \\    
36: \hspace{1cm} $Y_{min}$  $\gets$ $Y_{array}$ ${[}i{]}$                                         \\
37: \hspace{0.5cm} \textbf{else}                                                \\
38: \hspace{1cm} \textbf{if} $Y_{array}$ ${[}i{]}$ \textgreater $Y_{max}$ \hspace{0.1cm} \textbf{then}                           \\
39: \hspace{1.5cm} $Y_{max}$ $\gets$ $Y_{array}$ ${[}i{]}$                                 \\
40: \hspace{1cm} \textbf{end if}                                             \\
41: \hspace{0.5cm} \textbf{end if}                                              \\
42: \textbf{end for}                                             \\
43: $X_{position}$ \hspace{0.1cm} $\gets $ $X_{min}$ \hspace{1.3cm} $\to$ Update with final values \\            
44: $Y_{position}$  \hspace{0.15cm} $\gets $ $Y_{min}$   \\
45: $X_{size}$  \hspace{0.55cm} $\gets$ $X_{max} - X_{min}$  + 1 \\
46: $Y_{size}$ \hspace{0.59cm} $\gets$ $Y_{max}$ -  $Y_{min}$  + 1  \\ \hline
\end{tabular}
\end{table}

\subsection{Pre-Processing}
After getting the result of the Movement Detection procedure and getting a greater value than the threshold value, the algorithm will continue to execute the Pre-processing procedure. The processes of pre-processing consist of: 
\begin{enumerate}
\item Positioning the frame on a predetermined frame sample. Cropping the desired region from the inputted 3-channel image, After the ROI parameters are determined, the program will crop out the ROI with these parameters from the input BGR 3-channel video frame input. The kernel-internal input-output type for accelerated kernel arguments of Alveo is mostly stream-like data; therefore, the stream-architect implementation typed cropping will be used instead of the memory-mapped architecture implementation.
\item The cropping process will vary in size, so to input the cropped region into the neural network, resize it to the standard input size of the ImageNet model, which is 224 x 224 x 3, with the exception of the Inception-v4 model, whose input size is 299 x 299 x 3.
\item The process involves resizing the resulting image to a resolution of 224 x 224 and carrying out appropriate pre-processing tasks related to neural networks. The resize function offers a variety of interpolation settings, enabling the calculation of the intermediate point between two original pixels in the resized image. Interpolation settings include the nearest neighbor, bilinear, bicubic, area relation, etc. To preserve the 3-channel color feature while still enhancing performance, the result uses bilinear interpolation for the resizing function.
\item Before being fed for inference into a deep neural network, the image must be broken down into an array to be able to be fed into the inference graph of the neural network, in complement with the serialization step, which has several extra steps of pre-processing, depending on the neural network model.
\end{enumerate}

\subsection{CNN or Transformer Classifier}
This research employs four models with artificial neural networks and transformer models to recognize and classify objects and compare them to end-to-end methods. These models used the ImageNet dataset \cite{2015imagenet}, which is a large-scale visual database designed to advance research in computer vision and machine learning. The dataset contains over 14 million images, spanning more than 21,000 categories, with a subset of 1,000 categories often used for the ImageNet Large Scale Visual Recognition Challenge (ILSVRC). This subset includes approximately 1.2 million training images, 50,000 validation images, and 100,000 test images, making it a comprehensive resource for evaluating machine learning models. The specific models are ResNet50, MobileNet, Inception-v4, and ViT Base. 

\subsubsection{The architecture of MobileNet}
\begin{figure}[!h]
\centering
\includegraphics[width=0.4\textwidth]{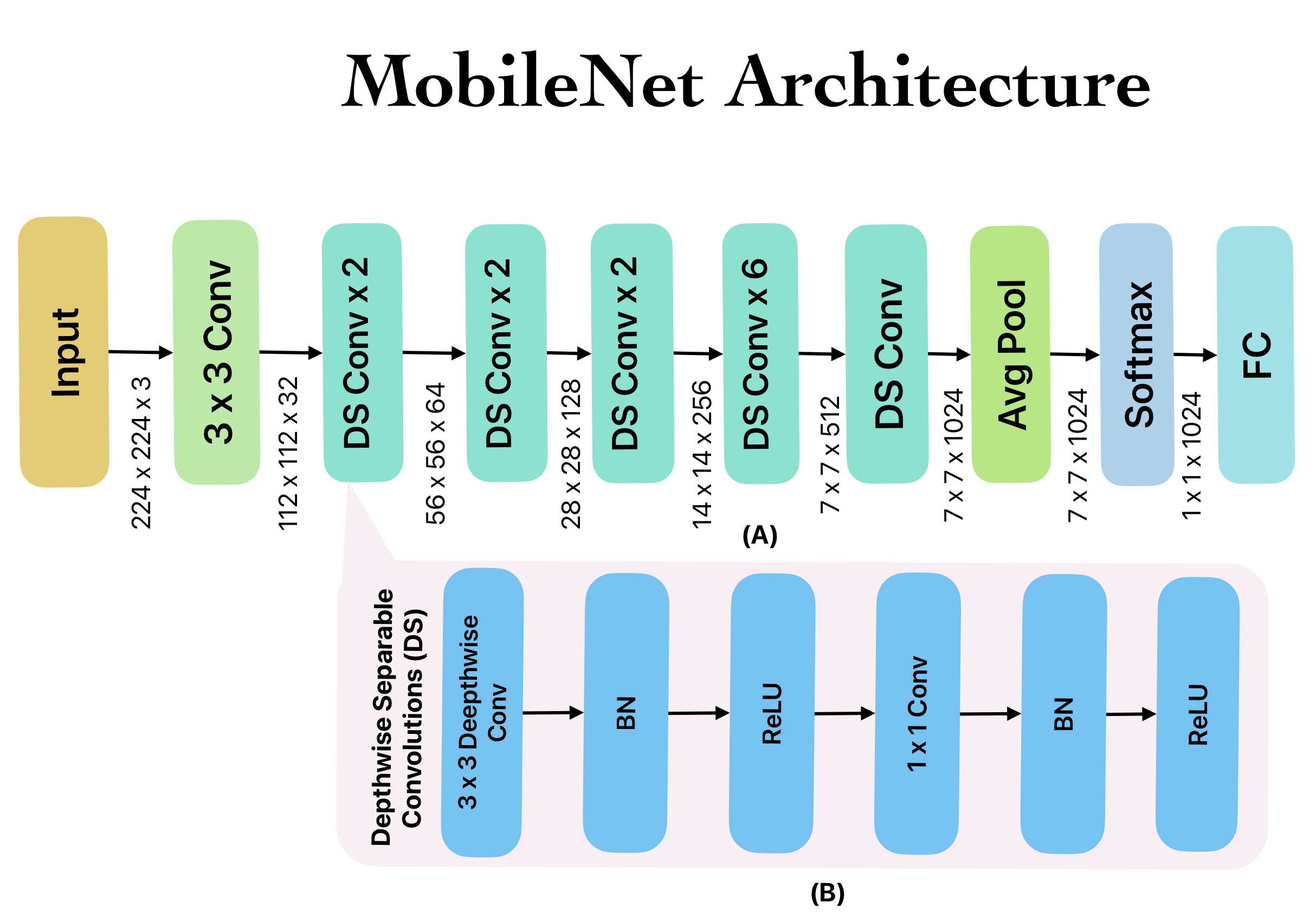}
\caption{The architecture Diagram of MobileNet}
\label{fig:mobilenet}
\end{figure}

MobileNet is specifically optimized for high efficiency and performance on mobile and embedded devices. Fig. \ref{fig:mobilenet} of the document illustrates the structure of the MobileNet architecture, highlighting its efficiency and modular design. Part (A) shows the overall architecture, which includes convolutional layers for feature extraction, depthwise separable convolutions (DS) to significantly reduce computational complexity, average pooling for spatial reduction, fully connected layers to map features to output classes, and a softmax layer for generating class probabilities. Part (B) provides a detailed view of the depthwise separable convolution process, which splits standard convolutions into two operations: depthwise convolution, which applies a filter to each input channel independently, and pointwise convolution (1x1), which combines these filtered outputs. The DS layer also incorporates batch normalization (BN) to stabilize and accelerate training and a rectified linear unit (ReLU) for non-linear activation, enhancing the network’s ability to model complex patterns.

\subsubsection{The architecture of Inception-V4}
\begin{figure*}[!h]
\centering
\includegraphics[width=0.7\textwidth]{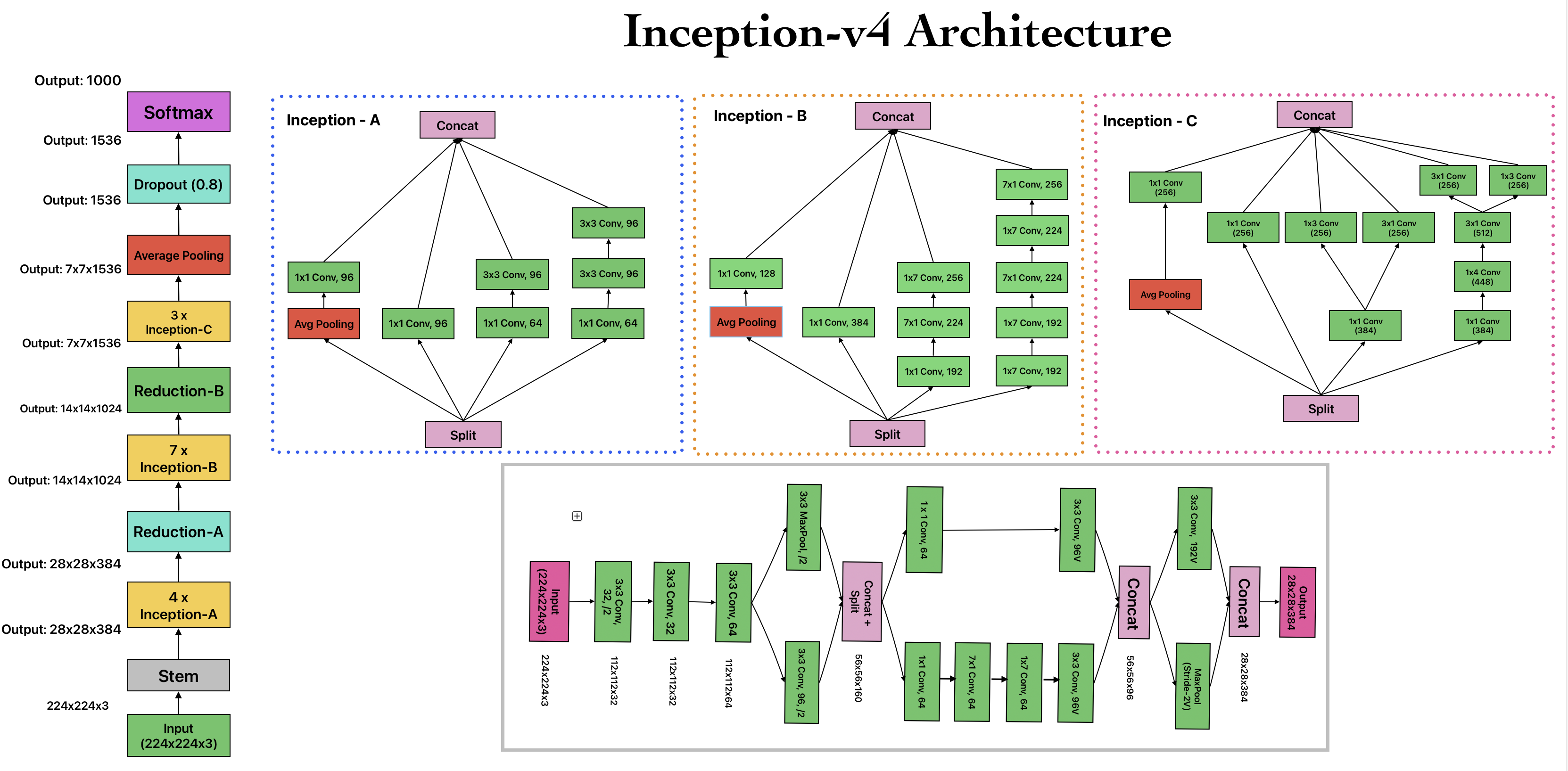}
\caption{The architecture Diagram of Inception-v4}
\label{fig:inception}
\end{figure*}

Inception V4 enhances the Inception architecture by incorporating an inception module that includes parallel convolutional layers of varying widths. In Fig. \ref{fig:inception} of the document, the architecture of the Inception-v4 model is illustrated. It includes several interconnected layers designed to optimize deep learning performance. The core components are the Inception blocks, which feature residual connections for efficient gradient flow and improved training stability. These blocks combine various filter sizes in parallel, enabling the model to extract features at multiple scales. The architecture integrates batch normalization layers to stabilize learning and reduce overfitting, and residual scaling is applied to prevent gradients from vanishing in deeper networks. Additionally, the model incorporates hyperparameter tuning for optimization and performance improvement, which is further enhanced by using Bayesian optimization. The diagram showcases the systematic arrangement of these components to achieve robust feature extraction and classification.

\subsubsection{The architecture of ResNet50}
ResNet50 employs residual learning with 50 layers, incorporating skip connections to facilitate the smooth passage of gradients throughout the network, thus mitigating the issue of vanishing gradients in deep networks. Fig. \ref{fig:vit} is the architecture of the ResNet-50 model, a deep convolutional neural network designed to address the degradation problem in training deep networks. The diagram highlights the use of residual blocks, which are the core innovation in ResNet. These blocks introduce shortcut connections that bypass one or more layers, facilitating efficient training and ensuring that deeper networks can achieve better performance without degradation. The figure differentiates between two types of shortcut connections: identity shortcuts, used when input and output dimensions are the same, and projection shortcuts, used to match differing dimensions. Downsampling is achieved between blocks with a stride of 2, enabling spatial reduction and efficient feature extraction. The structure demonstrates how ResNet-50 balances depth and computational efficiency, making it highly effective for tasks such as image classification.

\begin{figure*}[!h]
\centering
\includegraphics[width=0.7\textwidth]{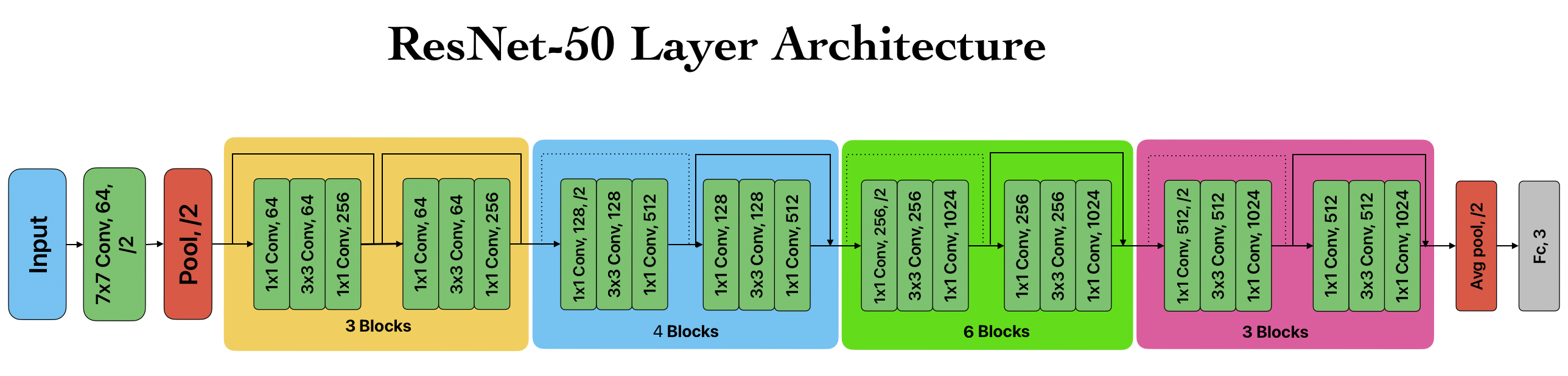}
\caption{The architecture Diagram of ResNet50}
\label{fig:resnet}
\end{figure*} 

\subsubsection{The architecture of Vision Transformers (ViT)}
\begin{figure}[!h]
\centering
\includegraphics[width=0.4\textwidth]{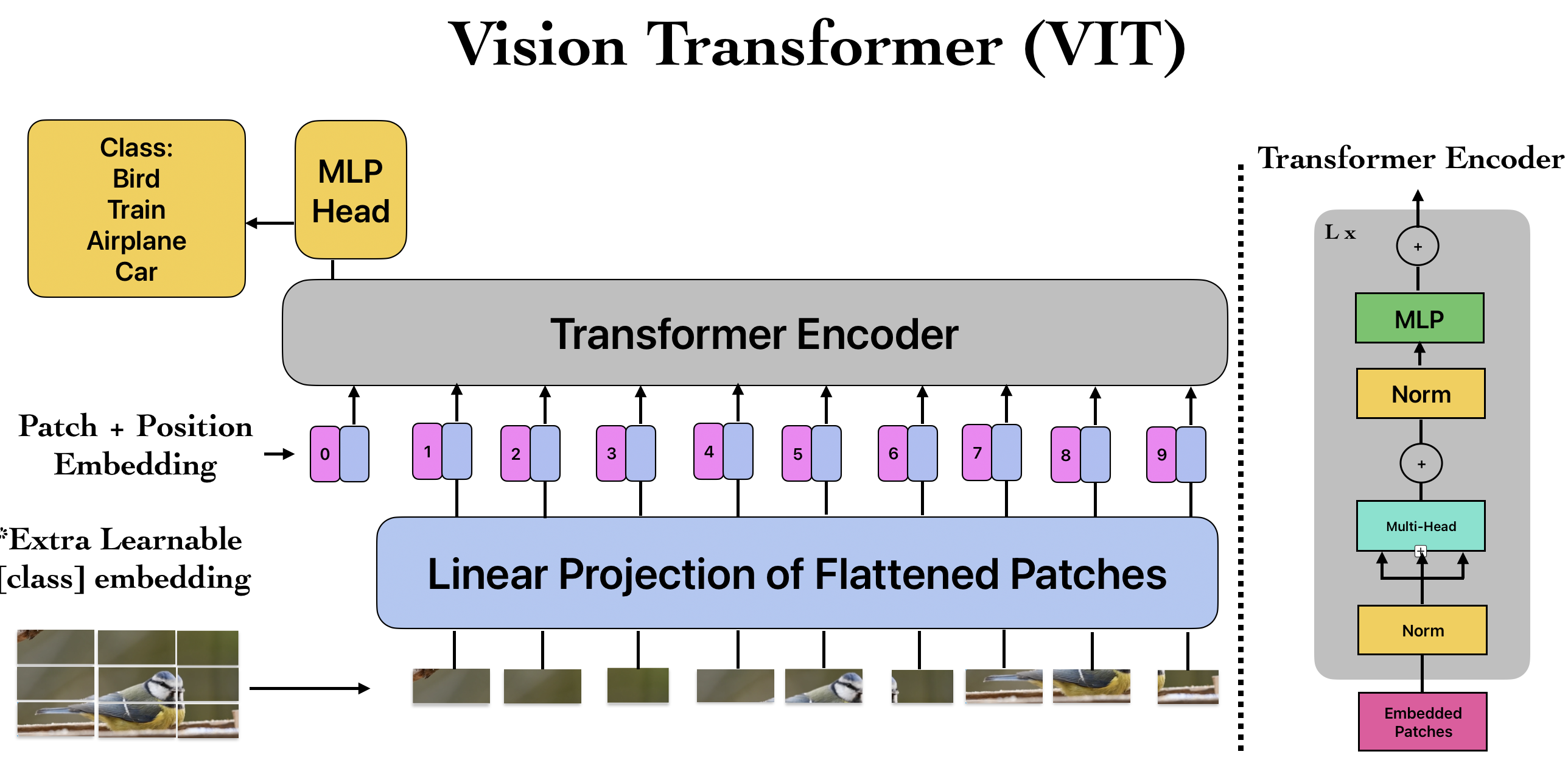}
\caption{The architecture Diagram of VIT}
\label{fig:vit}
\end{figure}

Vision Transformers (ViT), especially ViT Base, signify a transition from convolutional networks to transformer-based structures for visual applications. This model employs a self-attention technique to collect and incorporate global dependencies in input images effectively. The ViT Base model provides greater precision; albeit, this comes with the drawback of longer processing time and higher computational demands. Fig. \ref{fig:vit} is an overview of the Vision Transformer (ViT) architecture. It demonstrates how images are processed as sequences of patches rather than full 2D structures. The process begins by dividing the input image into fixed-size patches, flattening them, and applying a linear embedding to each patch to project it into a consistent vector space. Positional embeddings are added to these patch embeddings to encode spatial information, ensuring the model retains awareness of patch positions within the original image. These embeddings are then fed into a standard Transformer encoder, which comprises layers of multi-head self-attention and feed-forward networks, with residual connections and normalization applied throughout. This architecture leverages the scalability and effectiveness of transformers, originally designed for NLP tasks, for image recognition.

MobileNet is designed to be efficient and low-latency in resource-constrained environments. Inception V4 focuses on achieving high accuracy using more computational power. ResNet50 strikes a balance between depth and efficiency. The ViT model offers advanced accuracy with scalable complexity, making it suitable for applications that require a thorough understanding of global features. Each model has been trained using ImageNet as the base dataset. The reason for choosing MobileNet, Inception V4, ResNet50, and several Vision Transformers is that they are leading neural network designs developed to achieve various trade-offs between accuracy, latency, and efficiency in computer vision tasks.

\subsection{Hardware Deployment}
\subsubsection{The AMD Alveo U50 and Hailo AI Accelerator}

\begin{figure}[h!]
\centering
\includegraphics[width=0.45\textwidth]{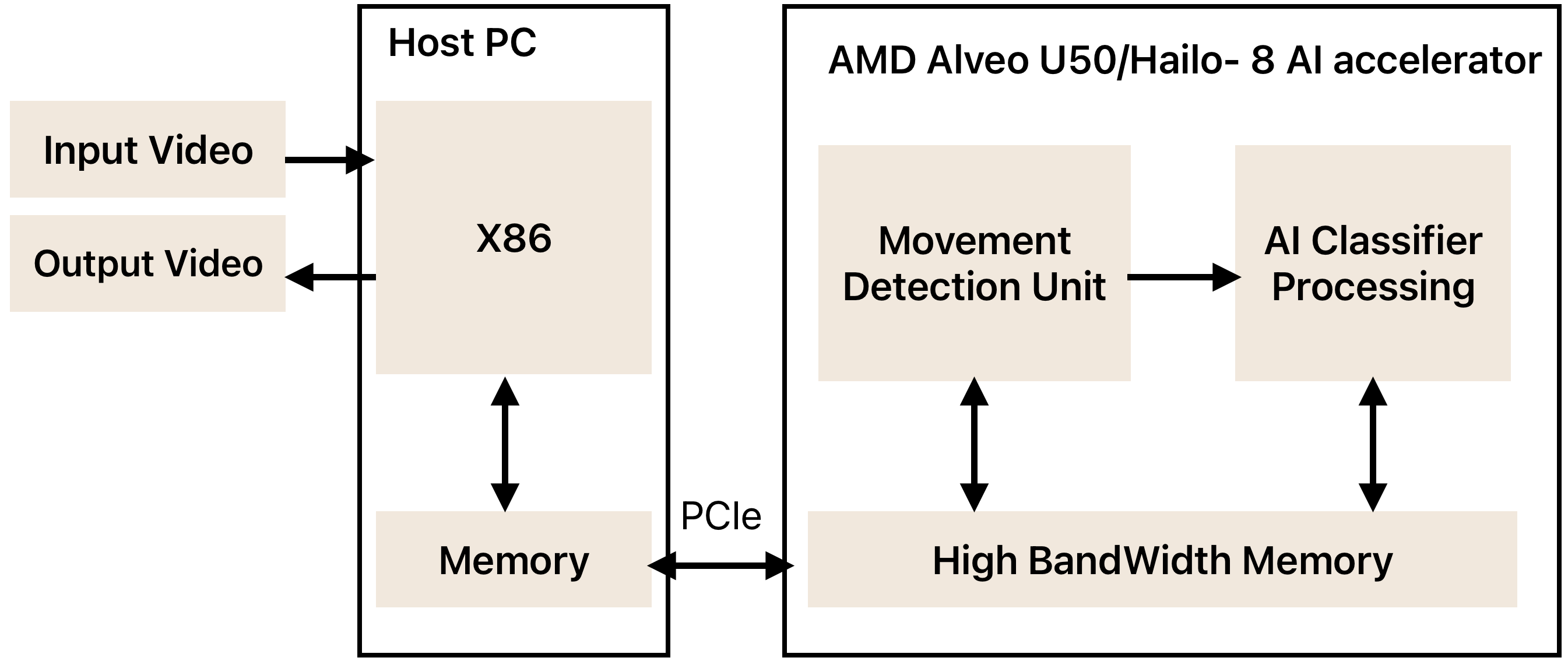}
\caption{The system architecture of Hailo and AMD Alveo ${TM}$ U50 }
\label{fig:hailofpga}
\end{figure}

Fig. \ref{fig:hailofpga} illustrates the structure of the object detection and classification system implemented on the Alveo U50 and Hailo-8 AI Accelerator. Both of these devices possess identical processing stages in terms of their architectural design. During the second processing phase, the system divides into two main components: the host PC and the device, specifically the AMD Alveo U50/Hailo-8 accelerator. The host PC is an X86 CPU, renowned for its robust performance and adaptability in managing diverse computer activities, which equips the host PC. The input video, the X86 CPU carries out preprocessing operations to prepare the video data for next processing. This step may involve analyzing the video stream, implementing fundamental improvements, or refining the data to ensure it is in an optimal state for subsequent processing by AI accelerators. In addition to the X86 processor, PC Host also utilizes its memory subsystem. Visual data is temporarily stored and buffered in the memory component during the initial processing step.

This ensures seamless data transmission and enables the processor to quickly retrieve critical data without experiencing noticeable delays. After the host PC finishes its initial processing, it transfers the data to either the Alveo U50 or Hailo-8 for additional processing. High-performance artificial intelligence (AI) applications, specifically on both devices, make them well-suited for demanding video processing workloads. The system comprises a motion detection unit and an AI classifier processing module, both of which closely interact with the high-bandwidth memory. The Motion Detection Unit is responsible for identifying and tracking motion within video frames. The device utilizes sophisticated algorithms to identify and examine motion, which is crucial for tasks such as surveillance, automated video editing, and activity recognition. The AMD Alveo U50 or FPGA-based designs are highly suitable for providing the substantial computational power and precision necessary for accurate motion detection. High-bandwidth memory stores and handles the data after motion detection. 

This form of memory is crucial for efficiently managing vast quantities of data at a rapid pace, guaranteeing that processing units can quickly retrieve the necessary data. High-bandwidth memory (HBM) enhances the speed of both reading and writing data, reduces delays, and offers the ability to analyze data in real time. The AI Classifier Processing module receives the video data once it detects motion. This module is responsible for executing a proposed algorithm model in order to categorize the identified items. The synergy among the Motion Detection Unit, AI Classifier Processing Module, and High Bandwidth Memory guarantees a seamless and effective data processing workflow. The AI classifier sends the analyzed video back to the host PC. Subsequently, the host PC does the conclusive processing task of categorizing the identified objects.

\subsubsection{NVIDIA Jetson Orin Nano}
\begin{figure}[!h]
\centering
\includegraphics[width=0.45\textwidth]{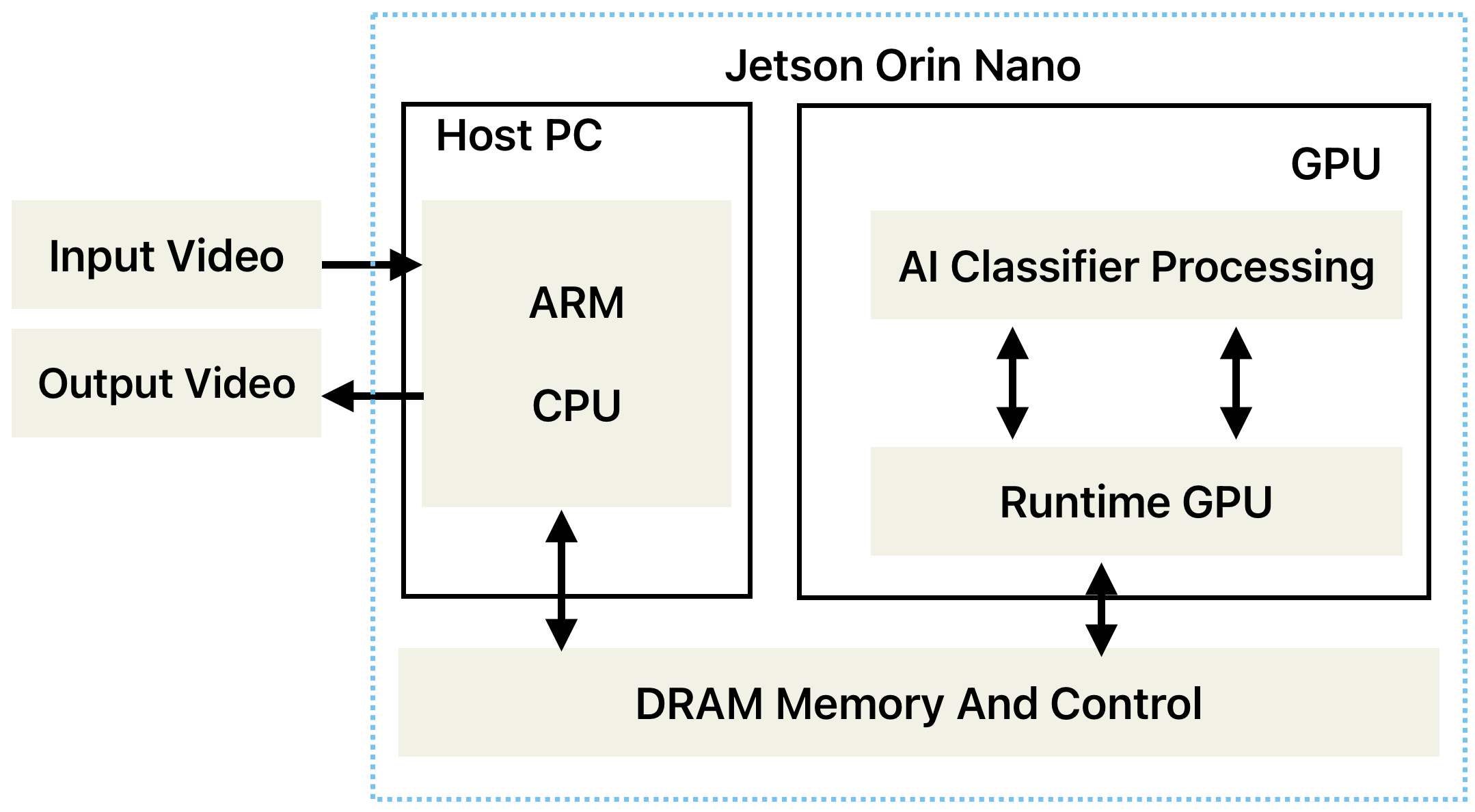}
\caption{The system architecture of Jetson Orin Nano}
\label{fig:jetson}
\end{figure}

The fig. \ref{fig:jetson} provides a detailed representation of the architecture of the Jetson Orin Nano-based video processing system for object detection. It illustrates the interconnections among different components to accomplish effective video processing and AI classification. The system comprises two primary components: the host PC and the GPU, both of which have significant roles in the whole workflow.The host PC receives the video input first. 

The ARM components and the host PC's CPU initially process the video. ARM processors specifically design themselves to efficiently handle fundamental processing tasks, offering a harmonious blend of performance and power efficiency. On the other hand, the CPU oversees complex computational processes, ensuring adequate preprocessing of the input video before it reaches the Jetson Orin Nano. The purpose of this pre-processing stage is to prepare the data for further AI processing.Once the host PC completes its initial processing, it transmits the data to the GPU via DRAM memory and control.GPUs are specialized modules specifically intended for AI and deep learning applications, making them well-suited for activities like real-time video processing and object detection. In addition, the memory and DRAM control modules have a significant impact. 

This module facilitates seamless data transmission across different system elements, offering the necessary bandwidth and memory administration for fast processing.The GPU is responsible for processing the Jetson Orin Nano's core. The GPU consists of two primary components: AI classifier processing and GPU runtime activities. Specifically designed for executing complex deep learning models and neural networks, the AI Classifier Processing component serves a variety of tasks such as object detection and other AI-driven video analysis activities. This component utilizes the parallel processing capabilities of GPUs to effectively manage the intricate calculations needed for contemporary AI algorithms.

The GPU Runtime section is responsible for overseeing real-time GPU operations. This entails performing the requisite computations and guaranteeing the timely delivery of the processed data. The interplay between the AI classifier processing and GPU runtime portions is crucial, as it enables the system to do complex AI tasks without sacrificing the real-time performance demanded by applications.After the GPU finishes its processing jobs, it produces a video output. The GPU then transmits the resulting video back to the host PC. The host PC receives the data and then performs the final processing of item detection and classification. The system's division of labor and efficient data handling make it well-suited for real-time video processing and AI applications, guaranteeing exceptional performance and dependability. The interactions among these components showcase the intricate nature and effectiveness of contemporary video processing systems, which are capable of flawlessly managing resource-intensive artificial intelligence tasks.

\section{EXPERIMENTAL RESULT}
This experiment uses four classes for object detection, such as birds, trains, airplanes, and cars. We are conducting experiments with our proposed method with four models, such as MobileNet \cite{mobilenet}, ResNet50 \cite{resnet}, Inception-v4 \cite{inception}, and ViT Base \cite{vit} using datasets from Imagenet \cite{2015imagenet}, and YOLOX \cite{yolox} using datasets from MS Coco \cite{2014coco}. Using some of the topologies, MobileNet, InceptionV4, ResNet50, ViT Base, and YOLOX. The tested data performance is accuracy (Acc\%), latency (ms), energy consumption (Joule), and efficiency (\%/mW). To calculate efficiency we use the formula from \cite{formula}. At Hailo-8 AI Accelerator and AMD Alveo U50, we use PowerTOP in the process of measuring latency, power, and energy during object detection tasks. PowerTOP is a Linux-based software developed by Intel to monitor the power consumption of computers or embedded systems. At Jetson Orin Nano we use jtop. Jtop is a monitoring tool specifically designed for NVIDIA Jetson devices, such as the Jetson Orin Nano. Same with PowerTOP, it can measure latency, power, and energy during object detection tasks.

We also consider the size of the video test we use. Then, we will implement it on three devices: the Hailo-8 AI Accelerator, the Jetson Orin Nano, and the AMD Alveo U50.

\begin{table*}[h!]
\centering
\caption{Evaluation of the Hailo - $8^{TM}$ AI Accelerator's performance}
\label{table:hai}
\begin{tabular}{c c c c c c c c}
\hline
%\rowcolor{gray}
\textbf{Class Name} & \textbf{Method}&\textbf{Topology} & \textbf{Acc(\%)} & \textbf{\begin{tabular}[c]{@{}c@{}}Latency\\(ms)\end{tabular}} & \textbf{\begin{tabular}[c]{@{}c@{}}Energy\\(Joule)\end{tabular}} & \textbf{\begin{tabular}[c]{@{}c@{}}Efficiency\\(\%/msW)\end{tabular}} & \textbf{Video Resolution}    \\ 
\hline 

& Proposed   & MobileNet & 92.6 &  35.63 & 1,221& 0.1731  &\\ 
&  Object Detection &InceptionV4 & 84.2 & 47.12 & 1,751 & 0.0541&  \\ 
Bird & Method  & ResNet50 & 72.3 & 49.16 & 1,184 & 0.0806 & 3840 x 2160\\ 
&     & ViT Base & 94.9 & 43.36 & 1,750 & 0.0875 &  \\ \hline
%\rowcolor{lightgray} 
Bird & Object Detection & YOLOX & 67.4 & 48.62 & 2,847 &  0.0393 & 3840 x 2160\\  \hline

& Proposed   & MobileNet & 97.7 & 43.80 & 4,849 &  0.046 & \\ 
&Object Detection & InceptionV4 & 57.4 & 65.23 & 5,892 & 0.025 & \\ 
Train&Method& ResNet50 & 95.7 & 56.71 & 5,094 & 0.045 & 4096 x 2016\\ 
& & ViT Base & 84.4 & 83.93 & 7.580 & 0.0370 & \\ \hline
%\rowcolor{lightgray}
Train& Object Detection & YOLOX & 57.4 & 49.54 & 10,880 &  0.0354  & 4096 x 2016 \\ \hline

& Proposed&MobileNet & 57.4 &  14.32 & 69.34 &  0.1168 & \\ 
&Object Detection & InceptionV4 & 44.4 & 16.85 & 67.41 & 0.1234  &\\ 
Airplane&Method& ResNet50 & 28.4 & 15.68 & 69.00 & 0.0631  &3840 x 2160\\ 
&  & ViT Base & 45.6 & 19.01 & 108.41 & 0.1644&  \\ \hline
%\rowcolor{lightgray} 
Airplane & Object Detection & YOLOX & 25.9 & 39.80 & 274.65 &  0.02762 & 3840 x 2160\\  \hline
& Proposed& MobileNet & 98.4 &  22.07 & 241.05 &  0.1168 & \\ 
&Object Detection & InceptionV4 & 94.2 & 33.74 & 907.39 & 0.1582  &\\ 
Car&Method & ResNet50 & 99.8 & 27.91 & 245.25 & 0.1825  &4096 x 2016\\ 
& & ViT Base & 97.7 & 60.10 & 308.98 & 0.1410 &  \\ \hline
%\rowcolor{lightgray}
Car & Object Detection & YOLOX & 93.0 & 49.80 & 794.90 &  0.08658 & 4096 x 2016\\  \hline
\end{tabular}
\end{table*}

\subsection{Hailo - $8^{TM}$ AI Accelerator's performance}
The table \ref{table:hai} presents an experimental result of the method when implemented in the Hailo-8™ AI Accelerator. The Birds class, when utilizing the proposed method, obtains an accuracy of 92.6\% with MobileNet. MobileNet also has the lowest latency of 35.63 ms and the lowest energy consumption of 1,221 J. As a result, it has the highest efficiency of 0.1731 at a resolution of 3840 x 2160. The ViT Base model attains the utmost accuracy of 94.9\% while maintaining a reasonable level of latency and energy usage. However, InceptionV4 and ResNet50 demonstrate increased latencies and energy consumptions, resulting in reduced efficiency. The Bird class for object detection in YOLOX demonstrates a precision of 67.4\% with a latency of 48.62 ms and energy consumption of 2,847 J, yielding an efficiency of 0.0393 at a resolution of 3840 x 2160. 

The Train class achieves the maximum accuracy (97.7\%) with a latency of 43.801 ms and an energy usage of 4,849 J with MobileNet. InceptionV4 and ResNet50 exhibit elevated energy usage and latencies, leading to reduced efficiency. The Train class for Object Detection using YOLOX achieves an accuracy of 57.4\% with a latency of 49.54 ms and energy consumption of 10,880 J. This results in an efficiency of 0.0354 at a resolution of 4096 x 2016. 

Then, the Airplane class obtains the maximum accuracy (57.4\%) with the lowest energy usage (69.34 J) and latency (14.32 ms) using MobileNet. Among all the models, ViT Base exhibits the most significant energy usage, amounting to 108.41 Joules, as well as the longest latency, at 19.01 ms. The YOLOX model in the Airplane class for object detection achieves an accuracy of 25.9\%, with a latency of 39.80 ms and energy consumption of 274.65 J. This results in an efficiency of 0.02762 at a resolution of 3840 x 2160. 

The Car class achieves the maximum accuracy (99.8\%) with a latency of 27.91 ms and an energy usage of 245.25 J with ResNet50. MobileNet exhibits a modest level of energy usage, specifically 241.05 Joules, and latency, specifically 22.07 ms. The Car class for Object Detection using YOLOX achieves a precision of 93.0\% with a response time of 49.80 ms and energy usage of 794.90 joules, resulting in an efficiency of 0.08658 at a resolution of 4096 x 2016. 

In general, the performance of the Hailo-8™ AI Accelerator varies greatly depending on the structure and objective, with MobileNet often being a good choice due to its balanced accuracy, latency, energy consumption, and efficiency. The combination of frame difference and classification typically achieves greater levels of accuracy compared to object detection. Additionally, MobileNet consistently demonstrates low latency and energy consumption across different classes, making it highly efficient. On the other hand, YOLOX, although it excels at object detection, has a tendency to exhibit increased energy consumption and latency.
\vskip -3.5\baselineskip plus -3fil

\begin{table*}[h!]
\centering
\caption{Evaluation of the jetson's performance}
\label{table:performance_jetson}
\begin{tabular}{c c c c c c c c }
\hline
%\rowcolor{gray}
\textbf{Class Name} & \textbf{Method}&\textbf{Topology} & \textbf{Acc(\%)} & \textbf{\begin{tabular}[c]{@{}c@{}}Latency\\(ms)\end{tabular}} & \textbf{\begin{tabular}[c]{@{}c@{}}Energy\\(Joule)\end{tabular}} & \textbf{\begin{tabular}[c]{@{}c@{}}Efficiency\\(\%/msW)\end{tabular}} & \textbf{Video Resolution}    \\ 
\hline 

&Proposed& MobileNet & 100 &  41.38 & 0.3937 & 0.8332  &\\ 
& Object Detection&InceptionV4 & 83.68 & 52.43 & 0.59436 & 0.5319&  \\ 
Bird& Method&ResNet50 & 75.99 & 61.57 & 0.4864 & 0.5239 & 3840 x 2160\\ 
&      & ViT Base & 93.72 & 63.58 & 0.53064 & 0.4466 &  \\ \hline
%\rowcolor{lightgray} 
Bird & Object Detection & YOLOX & 66.92 & 61.67 & 0.674 &  0.4203 & 3840 x 2160\\  \hline

&Proposed& MobileNet & 100 & 53.35 & 0.0596 &  0.6693 & \\ 
&Object Detection& InceptionV4 & 58.62 & 61.49 & 0.1094 & 0.3177 & \\ 
Train&Method& ResNet50 & 97.70 & 81.33 & 0.0875 & 0.3533 &4096 x 2016 \\ 
& & ViT Base & 87.36 & 203.5 & 0.0624 & 0.134 & \\ \hline
%\rowcolor{lightgray}
Train& Object Detection & YOLOX & 55.39 & 55.44 & 0.0857 &  0.3568  & 4096 x 2016 \\ \hline

&Proposed & MobileNet & 60.23 &  53.36 & 0.035 &  0.1044 & \\ 
& Object Detection& InceptionV4 & 49.99 & 82.31 & 0.090 & 0.0216  &\\ 
Airplane& Method & ResNet50 & 29.1 & 65.42 & 0.060 & 0.0267  &3840 x 2160\\ 
&  & ViT Base & 46.01 & 67.05 & 0.048 & 0.0581 & \\ 
%\rowcolor{lightgray} 
Airplane & Object Detection & YOLOX & 23.11 & 76.74 & 0.063 & 0.0137 & 3840 x 2160\\  \hline

&Proposed& MobileNet & 100 &  28.02 & 0.0807 &  1.151 & \\ 
& Object Detection& InceptionV4 & 100 & 32.79 & 0.2261 & 0.8712  &\\ 
Car & Method & ResNet50 & 100 & 33 & 0.136 & 0.8912  &4096 x 2016\\ 
& & ViT Base & 95.77 & 56.59 & 0.1432 & 0.4701 &  \\ \hline
%\rowcolor{lightgray}
Car & Object Detection & YOLOX & 91.11 & 45.48 & 0.193 &  0.6907 & 4096 x 2016\\  \hline

\end{tabular}
\end{table*}

\subsection{The jetson's performance}
The table \ref{table:performance_jetson} assesses the performance of the method when implemented in Jetson in multiple classes. The Bird class, MobileNet achieves a flawless accuracy of 100\%. It also has a low latency of 41.38 ms and consumes just 0.3937 J of energy. This makes MobileNet the most efficient option, with an efficiency of 0.8332\% /ms W. ViT Base demonstrates a remarkable accuracy of 93.72\% with reasonable latency and energy usage, whereas InceptionV4 and ResNet50 exhibit lesser accuracies and modest efficiency. For object detection in the Bird class, YOLOX demonstrates a precision of 66.92\% with a latency of 61.67 milliseconds and power usage of 0.674 joules, leading to an efficiency of 0.4203\% /ms W at a resolution of 3840 x 2160. 

The Trains class, using the proposed method, demonstrates MobileNet's consistent accuracy of 100\%. This is achieved with a latency of 53.35 ms and remarkably low energy consumption of 0.0596 J. ResNet50 achieves a high level of accuracy (97.70\%), but it also exhibits increased latency and energy consumption, and ViT Base demonstrates middling performance metrics. The Train class for Object Detection demonstrates that YOLOX attains a precision of 55.39\% with a latency of 55.44 ms and power consumption of 0.0857 J, yielding an efficiency of 0.3568\%/mW at a resolution of 4096 x 2016. 

The Airplanes class obtains an accuracy of 60.23\% with a latency of 53.36 ms and an energy usage of 0.035 J when employing MobileNet. ViT Base demonstrates the highest levels of latency and energy consumption, resulting in reduced efficiency. The Airplane class in Object Detection using YOLOX achieves an accuracy of 23.11\% with a latency of 76.74 ms and energy consumption of 0.063 J. This results in an efficiency of 0.0137 at a resolution of 3840 x 2160. 

The Car class achieves perfect accuracy (100\%) with a latency of 28.02 ms and energy usage of 0.0807 J, resulting in the maximum efficiency (1.151\%/mW) at a resolution of 4096 x 2016, as demonstrated by MobileNet. The Car class for Object Detection in YOLOX demonstrates a precision of 91.11\% with a response time of 45.48 ms and power consumption of 0.193 joules, yielding an efficiency of 0.6907 at a resolution of 4096 x 2016. The performance of Jetson's device is influenced by its structure and objectives, with MobileNet continuously demonstrating high accuracy, low latency, low power usage, and high energy efficiency across many classes. The combination of frame difference and classification typically achieves greater levels of accuracy compared to object detection. MobileNet, in particular, demonstrates remarkable performance in terms of both accuracy and efficiency. Although YOLOX has strong performance in object detection tasks, it is characterized by longer response times and worse operational efficiencies when compared to alternative topologies.

\begin{table*}[h!]
\centering

\caption{Evaluation of the The AMD Alveo U50's performance }

\label{table:performance_fpga1}
\begin{tabular}{c c c c c c c c}
\hline
%\rowcolor{gray}
\textbf{Class Name} & \textbf{Method}&\textbf{Topology} & \textbf{Acc(\%)} & \textbf{\begin{tabular}[c]{@{}c@{}}Latency\\(ms)\end{tabular}} & \textbf{\begin{tabular}[c]{@{}c@{}}Energy\\(Joule)\end{tabular}} & \textbf{\begin{tabular}[c]{@{}c@{}}Efficiency\\(\%/msW)\end{tabular}} & \textbf{Video Resolution}    \\ 
\hline 
& & MobileNet & 94.39 & 7.74 & 43.74 &  0.8935 & \\ 
Bird &Proposed Method& InceptionV4 & 85.02 & 12.82 & 111.12 & 0.220 &3840 x 2160  \\ 
  &   &ResNet50 & 77.68 & 11.55 & 95.46 & 0.2134 & \\ \hline
%\rowcolor{lightgray}
Bird& Object Detection & YOLOX & 69.11 & 16.37 & 347.75 &  0.1189 & 3840 x 2160\\ \hline

& & MobileNet & 93 & 8.44 & 30.91 &  0.989  & \\ 
Trains &Proposed Method& InceptionV4 & 54.52 & 12.23 & 120.24 & 0.1815 & 4096 x 2016\\ 
 &   & ResNet50 & 90.86 & 11.29 & 89.63 & 0.3129 &  \\ \hline
%\rowcolor{lightgray}
Train & Object Detection & YOLOX & 51.52 & 13.15 & 318.89 &  0.135 & 4096 x 2016  \\ \hline

&  & MobileNet & 59.32 & 8.39 & 25.12 &  0.6  &\\ 
Airplanes &Proposed Method& InceptionV4 & 51.67 & 12.68 & 97.72 & 0.157 &3840 x 2160 \\ 
 &   & ResNet50 & 31.11 & 9.08 & 72.84 & 0.1269 & \\\hline
%\rowcolor{lightgray} 
Airplane & Object Detection & YOLOX & 29.11 & 16.88 & 350.67 &  0.0580 & 3840 x 2160\\ \hline

& & MobileNet & 97.67 & 10.58 & 23.04 &  0.78103  & \\ 
Car &Proposed Method& InceptionV4 & 91.12 & 13.20 & 89.612 & 0.2902 & 4096 x 2016\\ 
 &   & ResNet50 & 93 & 13.06 & 66.797 & 0.2855 & \\ \hline
\%rowcolor{lightgray}
Car & Object Detection & YOLOX & 84.73 & 11.28 & 321.55 &  0.2673 & 4096 x 2016 \\ \hline

\end{tabular}
\end{table*}

\subsection{The AMD Alveo U50's performance}
The tables \ref{table:performance_fpga1} show the results of how well the method was implemented in an AMD Alveo ${TM}$ U50 and could identify four different types of objects: birds, trains, airplanes, and cars. The ViT model’s performance evaluation results are not presently accessible on Alveo devices because Vitis AI does not currently support it. Therefore, testing is restricted to four models for Alveo devices: MobileNet, InceptionV4, Resnet50, and YOLOX.

For the Bird class, MobileNet achieves high accuracy (94.39\%) with low latency (7.74 ms) and energy consumption (43.74 J), leading to the highest efficiency (0.8935\%/mW) at a resolution of 3840 x 2160. InceptionV4 and ResNet50 show moderate accuracies and latencies with higher energy consumptions, resulting in lower efficiencies. In the Bird class for Object Detection, YOLOX achieves an accuracy of 69.11\% with a latency of 16.37 ms and energy consumption of 347.75 J, resulting in an efficiency of 0.1189 at a resolution of 3840 x 2160. 

For the Trains class, MobileNet shows high accuracy (93\%) with low latency (8.44 ms) and minimal energy consumption (30.91 J). ResNet50 follows with a high accuracy (90.86\%) but slightly higher latency and energy consumption. InceptionV4 demonstrates lower accuracy and efficiency compared to MobileNet and ResNet50. In the Train class for Object Detection, YOLOX achieves an accuracy of 51.52\% with a latency of 13.15 ms and energy consumption of 318.89 J, resulting in an efficiency of 0.135 at a resolution of 4096 x 2016. 

For the Airplanes class, MobileNet achieves moderate accuracy (59.32\%) with low latency (8.39 ms) and energy consumption (25.12 J). InceptionV4 and ResNet50 show lower accuracies and efficiencies, with higher latencies and energy consumptions. In the Airplane class for Object Detection, YOLOX achieves an accuracy of 29.11\% with a latency of 16.88 ms and energy consumption of 350.67 J, resulting in an efficiency of 0.058 at a resolution of 3840 x 2160. 

For the Car class, MobileNet shows high accuracy (97.67\%) with moderate latency (10.581 ms) and low energy consumption (23.04 J), leading to high efficiency (0.78103\%/mW). ResNet50 and InceptionV4 also demonstrate high accuracies with slightly higher latencies and energy consumptions. In the Car class for Object Detection, YOLOX achieves an accuracy of 84.73\% with a latency of 11.28 ms and energy consumption of 321.55 J, resulting in an efficiency of 0.2673\%/mW at a resolution of 4096 x 2016. So, the Alveo's performance varies across different topologies and tasks, with MobileNet consistently demonstrating high accuracy, low latency, low energy consumption, and high efficiency across various classes. 

Frame difference + classification generally yields higher accuracies and efficiencies than object detection. YOLOX, while performing well in object detection, tends to have higher latencies and lower efficiencies compared to other topologies. Overall, MobileNet appears to be the most balanced option for Alveo, providing high performance across multiple metrics. So, from all the experimental studies that we examined in the various classes, we conclude that using our proposed method, the MobileNet model has consistently high accuracy, low latency, and highly efficient energy for all the devices that we tested. Otherwise, YOLOX consistently demonstrates the lowest accuracy, the lowest latency, and the lowest efficiency. If we compare our proposed algorithm and end-to-end method, our method increases average accuracy by 28.314\%, an average efficiency of 3.6 times, and an average latency reduction of 39.305\% compared to the end-to-end method. 

Frame differencing can be more efficient because it utilizes localized operations based on pixel changes, while end-to-end methods such as YOLO process global features of the entire image, requiring more computation and thus needing higher energy consumption. MobileNet, ResNet50, ViT Base, and InceptionV4 each have strengths and shortcomings in fast object detection tasks using the frame difference method. MobileNet demonstrates high accuracy, low latency, and exceptional energy efficiency, making it ideal for IoT systems and edge devices, but its lightweight architecture may struggle with highly complex detection tasks. ResNet50 achieves a good balance of accuracy and efficiency but has slightly higher latency and energy consumption than MobileNet, limiting its suitability for resource-constrained environments. ViT Base excels in accuracy due to its transformer-based architecture, which captures global context effectively but suffers from high latency and significant energy demands and is less suited for real-time or energy-sensitive applications. InceptionV4 delivers high accuracy and robust detection but has the highest latency and energy consumption among the models, making it unsuitable for time-critical or energy-constrained tasks. YOLOX exhibits lower accuracy, higher latency, and less energy efficiency compared to methods utilizing the frame difference approach, as the end-to-end methodology employed by YOLOX struggles to handle the computational and efficiency demands of fast-moving object detection, which are better addressed by lightweight algorithms like the frame difference method.

Some of the reasons why our method cannot achieve maximum accuracy are Frame differencing is very sensitive to environmental noise. In dynamic scenes where the background is not static, such as areas with moving trees, flowing water, or fluctuating lighting conditions, even small changes in the background can be detected as motion. This leads to a large number of false positives. In addition, any vibration or small shift in the camera position is interpreted as a difference between frames, which further increases false detections. 

Frame differencing has difficulty detecting motion in certain scenarios, such as small or slow objects. When objects move very slowly, the difference in pixel intensity between consecutive frames may be too subtle to pass the detection threshold, causing the method to miss these objects altogether. and very fast-moving objects until motion blur occurs. This blurring reduces the contrast and clarity of object edges, making it difficult for the frame differencing method to effectively identify and quantify changes.

\section{Conclusion}
This study presents an improved fast-moving object detection using the frame difference method. The proposed method will offer significant advantages in implementing fast-moving object identification in diverse sectors, such as surveillance, ADAS, and human activity recognition, with high computational speed and power efficiency. The implementation of our method used four object detection classes, such as birds, trains, airplanes, and cars, and some of the topologies were MobileNet, InceptionV4, ResNet50, and ViT Base. The tested data performance is accuracy (Acc\%), latency (ms), energy consumption (Joule), and efficiency (\%/msW). This proposed method has been implemented on the Alveo U50, Jetson Orin Nano, and Hailo-8 AI Accelerator. The result will compare our performance method to the YOLOX with the end-to-end method. 

The experimental results indicate that our method consistently demonstrated high efficiency, accuracy, and latency for the three devices that were examined. On the other hand, the end-to-end method consistently demonstrated the lowest level of accuracy, the most significant latency, and the poorest level of efficiency on devices in all parameters. Compared to the end-to-end method, the proposed method achieves an average latency efficiency gain of 3.6 times that of the YOLOX method. Additionally, it shows an average accuracy improvement of 28.3142\%. Compared to the YOLOX, the average latency reduction is 39.305\%. From the results, we conclude that the faster the movement of an object, the lower the accuracy will be obtained by the YOLOX method. Of all these classes, we know the fastest objects are trains and airplanes, and the percentage of accuracy in trains and airplanes is lower than in other categories. 

End-to-end object detection methods, such as YOLO (You Only Look Once), consume more energy than simpler techniques like frame differencing due to their computational complexity and resource demands. YOLO performs tasks like feature extraction, classification, and bounding box regression in a single pass. In contrast, frame differencing is a straightforward method that detects motion by comparing pixel values between consecutive frames. It doesn’t require complex computations or large-scale data processing, resulting in significantly lower energy consumption. 

This method innovatively enhances motion detection and object classification in energy-constrained IoT applications. Unlike traditional frame difference techniques, which rely solely on pixel-based motion detection, this method leverages AI classification to improve detection accuracy and efficiency. This hybrid approach enables real-time object recognition with higher accuracy, reduced latency, and optimized energy consumption, making it different from conventional frame difference methods and more suitable for the dynamic requirements of fast-moving object detection. So, our proposed method is a lightweight detection algorithm that is well-suited for detecting fast-moving objects and has higher accuracy.
\bibliographystyle{ieeetr}

\bibliography{bibli}

@inproceedings{formula,
  title={Edgevits: Competing light-weight cnns on mobile devices with vision transformers},
  author={Pan, Junting and Bulat, Adrian and Tan, Fuwen and Zhu, Xiatian and Dudziak, Lukasz and Li, Hongsheng and Tzimiropoulos, Georgios and Martinez, Brais},
  booktitle={European Conference on Computer Vision},
  pages={294--311},
  year={2022},
  organization={Springer}
}

@article{RR3,
  title={Service function chain embedding meets machine learning: Deep reinforcement learning approach},
  author={Liu, Yicen and Zhang, Junning},
  journal={IEEE Transactions on Network and Service Management},
  year={2024},
  publisher={IEEE}
}

@article{RR2,
  title={Pain without gain: Destructive beamforming from a malicious RIS perspective in IoT networks},
  author={Lin, Zhi and Niu, Hehao and An, Kang and Hu, Yihua and Li, Dong and Wang, Jiangzhou and Al-Dhahir, Naofal},
  journal={IEEE Internet of Things Journal},
  year={2023},
  publisher={IEEE}
}

@article{yolox,
author = {Zhang, Jianfei and Ke, Sai},
year = {2022},
month = {03},
pages = {1-8},
title = {Improved YOLOX Fire Scenario Detection Method},
volume = {2022},
journal = {Wireless Communications and Mobile Computing},
doi = {10.1155/2022/9666265}
}

@article{vit,
  title={An image is worth 16x16 words: Transformers for image recognition at scale},
  author={Alexey, Dosovitskiy},
  journal={arXiv preprint arXiv: 2010.11929},
  year={2020}
}

@article{inception,
  title={Hyperparameter tuning deep learning for diabetic retinopathy fundus image classification},
  author={Shankar, K and Zhang, Yizhuo and Liu, Yiwei and Wu, Ling and Chen, Chi-Hua},
  journal={IEEE access},
  volume={8},
  pages={118164--118173},
  year={2020},
  publisher={IEEE}
}

@inproceedings{mobilenet,
  title={Food image classification with improved MobileNet architecture and data augmentation},
  author={Phiphiphatphaisit, Sirawan and Surinta, Olarik},
  booktitle={Proceedings of the 3rd International Conference on Information Science and Systems},
  pages={51--56},
  year={2020}
}

@article{resnet,
  title={A classification of Arab ethnicity based on face image using deep learning approach},
  author={Al-Humaidan, Norah A and Prince, Master},
  journal={IEEE Access},
  volume={9},
  pages={50755--50766},
  year={2021},
  publisher={IEEE}
}

@inproceedings{2014coco,
  title={Microsoft coco: Common objects in context},
  author={Lin, Tsung-Yi and Maire, Michael and Belongie, Serge and Hays, James and Perona, Pietro and Ramanan, Deva and Doll{\'a}r, Piotr and Zitnick, C Lawrence},
  booktitle={Computer Vision--ECCV 2014: 13th European Conference, Zurich, Switzerland, September 6-12, 2014, Proceedings, Part V 13},
  pages={740--755},
  year={2014},
  organization={Springer}
}

@article{2015imagenet,
  title={Imagenet large scale visual recognition challenge},
  author={Russakovsky, Olga and Deng, Jia and Su, Hao and Krause, Jonathan and Satheesh, Sanjeev and Ma, Sean and Huang, Zhiheng and Karpathy, Andrej and Khosla, Aditya and Bernstein, Michael and others},
  journal={International journal of computer vision},
  volume={115},
  pages={211--252},
  year={2015},
  publisher={Springer}
}

@article{nov1,
  title={BSFD: BACKGROUND SUBTRACTION FRAME DIFFERENCE ALGORITHM FOR MOVING OBJECT DETECTION AND EXTRACTION.},
  author={Alex, D Stalin and Wahi, Amitabh},
  journal={Journal of Theoretical \& Applied Information Technology},
  volume={60},
  number={3},
  year={2014}
}

@inproceedings{nov2,
  title={Motion Detection in Real-Time Surveillance Using Two Frame Differencing},
  author={Biswas, Tamal and Bhattacharya, Diptendu and Rudrapal, Dwijen and Roy, Srijan and Mandal, Gouranga},
  booktitle={International Conference on Information and Communication Technology for Competitive Strategies},
  pages={97--109},
  year={2023},
  organization={Springer}
}

@article{kk4,
  title={Motion detection based on frame difference method},
  author={Singla, Nishu},
  journal={International Journal of Information \& Computation Technology},
  volume={4},
  number={15},
  pages={1559--1565},
  year={2014}
}

@article{nov3,
  title={Moving object detection based on frame difference and W4},
  author={Sengar, Sandeep Singh and Mukhopadhyay, Susanta},
  journal={Signal, Image and Video Processing},
  volume={11},
  pages={1357--1364},
  year={2017},
  publisher={Springer}
}

@ARTICLE{new1,
  author={Zhu, Sha and Ota, Kaoru and Dong, Mianxiong},
  journal={IEEE Internet of Things Journal}, 
  title={Energy-Efficient Artificial Intelligence of Things With Intelligent Edge}, 
  year={2022},
  volume={9},
  number={10},
  pages={7525-7532},
  doi={10.1109/JIOT.2022.3143722}}

@ARTICLE{new2,
  author={Guo, Siyan and Zhao, Cong and Wang, Guiqin and Yang, Jiaqing and Yang, Shusen},
  journal={IEEE Internet of Things Journal}, 
  title={EC²Detect: Real-Time Online Video Object Detection in Edge-Cloud Collaborative IoT}, 
  year={2022},
  volume={9},
  number={20},
  pages={20382-20392},
  doi={10.1109/JIOT.2022.3173685}}

@ARTICLE{r1,
  author={Chen, Chang Wen},
  journal={IEEE Internet of Things Journal}, 
  title={Internet of Video Things: Next-Generation IoT With Visual Sensors}, 
  year={2020},
  volume={7},
  number={8},
  pages={6676-6685},
  doi={10.1109/JIOT.2020.3005727}}

@ARTICLE{r22,
  author={Suo, Jiashun and Zhang, Xingzhou and Shi, Weisong and Zhou, Wei},
  journal={IEEE Internet of Things Journal}, 
  title={E3-UAV: An Edge-Based Energy-Efficient Object Detection System for Unmanned Aerial Vehicles}, 
  year={2024},
  volume={11},
  number={3},
  pages={4398-4413},
  doi={10.1109/JIOT.2023.3301623}}

@ARTICLE{reff4,
  author={Liang, Chung-Wei and Juang, Chia-Feng},
  journal={IEEE Transactions on Intelligent Transportation Systems}, 
  title={Moving Object Classification Using a Combination of Static Appearance Features and Spatial and Temporal Entropy Values of Optical Flows}, 
  year={2015},
  volume={16},
  number={6},
  pages={3453-3464},
  doi={10.1109/TITS.2015.2459917},
  ISSN={1558-0016},
  month={Dec},}

@article{reff6,
title = {Detection of moving objects based on enhancement of optical flow},
journal = {Optik},
volume = {145},
pages = {130-141},
year = {2017},
issn = {0030-4026},
doi = {https://doi.org/10.1016/j.ijleo.2017.07.040},
url = {https://www.sciencedirect.com/science/article/pii/S0030402617308574},
author = {Sandeep Singh Sengar and Susanta Mukhopadhyay}
}

@article{reff7,
  title={An Object Detection Method Using Wavelet Optical Flow and Hybrid Linear-Nonlinear Classifier},
  author={Pengcheng Han and Junping Du and Jun Zhou and Suguo Zhu},
  journal={Mathematical Problems in Engineering},
  year={2013},
  volume={2013},
  pages={1-14},
  url={https://api.semanticscholar.org/CorpusID:122017371}
}

@inproceedings{reff8,
author = {Lucas, Bruce D. and Kanade, Takeo},
title = {An iterative image registration technique with an application to stereo vision},
year = {1981},
publisher = {Morgan Kaufmann Publishers Inc.},
address = {San Francisco, CA, USA},
booktitle = {Proceedings of the 7th International Joint Conference on Artificial Intelligence - Volume 2},
pages = {674–679},
numpages = {6},
location = {Vancouver, BC, Canada},
series = {IJCAI'81}
}

@article{reff9,
title = {Determining optical flow},
journal = {Artificial Intelligence},
volume = {17},
number = {1},
pages = {185-203},
year = {1981},
issn = {0004-3702},
doi = {https://doi.org/10.1016/0004-3702(81)90024-2},
url = {https://www.sciencedirect.com/science/article/pii/0004370281900242},
author = {Berthold K.P. Horn and Brian G. Schunck}
}

@article{reff12,
  title={Fast detection of moving object based on improved frame-difference method},
  author={Ming Zhu and Hongbo Wang},
  journal={2017 6th International Conference on Computer Science and Network Technology (ICCSNT)},
  year={2017},
  pages={299-303},
  url={https://api.semanticscholar.org/CorpusID:5048876}
}

@inproceedings{reff13,
author = {Tsai, Chun-Ming and Yeh, Zong-Mu},
title = {Intelligent moving objects detection via adaptive frame differencing method},
year = {2013},
isbn = {9783642365454},
publisher = {Springer-Verlag},
address = {Berlin, Heidelberg},
url = {https://doi.org/10.1007/978-3-642-36546-1_1},
doi = {10.1007/978-3-642-36546-1_1},
booktitle = {Proceedings of the 5th Asian Conference on Intelligent Information and Database Systems - Volume Part I},
pages = {1–11},
numpages = {11},
location = {Kuala Lumpur, Malaysia},
series = {ACIIDS'13}
}

@INPROCEEDINGS{reff14,
  author={Jusman, Yessi and Hinggis, Lentera and Wiyagi, Rama Okta and Isa, Nor Ashidi Mat and Mujaahid, Faaris},
  booktitle={2020 1st International Conference on Information Technology, Advanced Mechanical and Electrical Engineering (ICITAMEE)}, 
  title={Comparison of Background Subtraction and Frame Differencing Methods for Indoor Moving Object Detection}, 
  year={2020},
  volume={},
  number={},
  pages={214-219},
  doi={10.1109/ICITAMEE50454.2020.9398484}}

@article{kk2,
  title={Dpsnet: Multitask learning using geometry reasoning for scene depth and semantics},
  author={Zhang, Junning and Su, Qunxing and Tang, Bo and Wang, Cheng and Li, Yining},
  journal={IEEE Transactions on Neural Networks and Learning Systems},
  volume={34},
  number={6},
  pages={2710--2721},
  year={2021},
  publisher={IEEE}
}

@article{kk1,
  title={Supporting IoT with rate-splitting multiple access in satellite and aerial-integrated networks},
  author={Lin, Zhi and Lin, Min and De Cola, Tomaso and Wang, Jun-Bo and Zhu, Wei-Ping and Cheng, Julian},
  journal={IEEE Internet of Things Journal},
  volume={8},
  number={14},
  pages={11123--11134},
  year={2021},
  publisher={IEEE}
}

@article{ref17,
  title={Learning a convolutional neural network for non-uniform motion blur removal},
  author={Jian Sun and Wenfei Cao and Zongben Xu and Jean Ponce},
  journal={2015 IEEE Conference on Computer Vision and Pattern Recognition (CVPR)},
  year={2015},
  pages={769-777},
  url={https://api.semanticscholar.org/CorpusID:1485453}
}

@article{reff18,
title = {Suitability of recent hardware accelerators (DSPs, FPGAs, and GPUs) for computer vision and image processing algorithms},
journal = {Signal Processing: Image Communication},
volume = {68},
pages = {101-119},
year = {2018},
issn = {0923-5965},
doi = {https://doi.org/10.1016/j.image.2018.07.007},
url = {https://www.sciencedirect.com/science/article/pii/S0923596518303606},
author = {Amir HajiRassouliha and Andrew J. Taberner and Martyn P. Nash and Poul M.F. Nielsen}
}

@article{reff20,
title = {Medical image processing on the GPU – Past, present and future},
journal = {Medical Image Analysis},
volume = {17},
number = {8},
pages = {1073-1094},
year = {2013},
issn = {1361-8415},
doi = {https://doi.org/10.1016/j.media.2013.05.008},
url = {https://www.sciencedirect.com/science/article/pii/S1361841513000820},
author = {Anders Eklund and Paul Dufort and Daniel Forsberg and Stephen M. LaConte}
}

@article{ref21,
author = {şimşek, Emrah and Ozyer, Baris},
year = {2021},
month = {05},
pages = {48-54},
title = {Selected Three Frame Difference Method for Moving Object Detection},
volume = {9},
journal = {International Journal of Intelligent Systems and Applications in Engineering},
doi = {10.18201/ijisae.2021.233}
}

@article{11,
  title={ImageNet classification with deep convolutional neural networks},
  author={Krizhevsky, Alex and Sutskever, Ilya and Hinton, Geoffrey E},
  journal={Communications of the ACM},
  volume={60},
  number={6},
  pages={84--90},
  year={2017},
  publisher={AcM New York, NY, USA}
}

@article{12,
  title={Novel convolutional neural network-based roadside unit for accurate pedestrian localisation},
  author={Ojala, Risto and Veps{\"a}l{\"a}inen, Jari and Hanhirova, Jussi and Hirvisalo, Vesa and Tammi, Kari},
  journal={IEEE Transactions on Intelligent Transportation Systems},
  volume={21},
  number={9},
  pages={3756--3765},
  year={2019},
  publisher={IEEE}
}

@article{13,
  title={Gc-net: Gridding and clustering for traffic object detection with roadside lidar},
  author={Zhang, Liwen and Zheng, Jianying and Sun, Rongchuan and Tao, Yanyun},
  journal={IEEE Intelligent Systems},
  volume={36},
  number={4},
  pages={104--113},
  year={2020},
  publisher={IEEE}
}

@article{21,
  title={Faster r-cnn: Towards real-time object detection with region proposal networks},
  author={Ren, Shaoqing and He, Kaiming and Girshick, Ross and Sun, Jian},
  journal={Advances in neural information processing systems},
  volume={28},
  year={2015}
}

@inproceedings{22,
  title={Accelerating binarized convolutional neural networks with software-programmable FPGAs},
  author={Zhao, Ritchie and Song, Weinan and Zhang, Wentao and Xing, Tianwei and Lin, Jeng-Hau and Srivastava, Mani and Gupta, Rajesh and Zhang, Zhiru},
  booktitle={Proceedings of the 2017 ACM/SIGDA International Symposium on Field-Programmable Gate Arrays},
  pages={15--24},
  year={2017}
}

@article{23,
  title={Automatic bird species detection from crowd sourced videos},
  author={Li, Wen and Song, Dezhen},
  journal={IEEE Transactions on Automation Science and Engineering},
  volume={11},
  number={2},
  pages={348--358},
  year={2013},
  publisher={IEEE}
}

@inproceedings{24,
  title={Learning the state space based on flying pattern for bird detection},
  author={Tian, Shuman and Cao, Xianbin and Zhang, Baochang and Ding, Yuxin},
  booktitle={2017 Integrated Communications, Navigation and Surveillance Conference (ICNS)},
  pages={5B3--1},
  year={2017},
  organization={IEEE}
}

@inproceedings{26,
  title={Training deeper models by GPU memory optimization on TensorFlow},
  author={Meng, Chen and Sun, Minmin and Yang, Jun and Qiu, Minghui and Gu, Yang},
  booktitle={Proc. of ML Systems Workshop in NIPS},
  volume={7},
  year={2017}
}

@article{27,
  title={DaDianNao: A neural network supercomputer},
  author={Luo, Tao and Liu, Shaoli and Li, Ling and Wang, Yuqing and Zhang, Shijin and Chen, Tianshi and Xu, Zhiwei and Temam, Olivier and Chen, Yunji},
  journal={IEEE Transactions on Computers},
  volume={66},
  number={1},
  pages={73--88},
  year={2016},
  publisher={IEEE}
}

@inproceedings{28,
  title={Efficient and effective sparse LSTM on FPGA with bank-balanced sparsity},
  author={Cao, Shijie and Zhang, Chen and Yao, Zhuliang and Xiao, Wencong and Nie, Lanshun and Zhan, Dechen and Liu, Yunxin and Wu, Ming and Zhang, Lintao},
  booktitle={Proceedings of the 2019 ACM/SIGDA International Symposium on Field-Programmable Gate Arrays},
  pages={63--72},
  year={2019}
}

@article{29,
  title={DLAU: A scalable deep learning accelerator unit on FPGA},
  author={Wang, Chao and Gong, Lei and Yu, Qi and Li, Xi and Xie, Yuan and Zhou, Xuehai},
  journal={IEEE Transactions on Computer-Aided Design of Integrated Circuits and Systems},
  volume={36},
  number={3},
  pages={513--517},
  year={2016},
  publisher={IEEE}
}

@article{30,
  title={Large-scale FPGA-based convolutional networks},
  author={Farabet, Cl{\'e}ment and LeCun, Yann and Kavukcuoglu, Koray and Culurciello, Eugenio and Martini, Berin and Akselrod, Polina and Talay, Selcuk},
  journal={Scaling up machine learning: parallel and distributed approaches},
  volume={13},
  number={3},
  pages={399--419},
  year={2011}
}

@article{31,
  title={Data and hardware efficient design for convolutional neural network},
  author={Lin, Yue-Jin and Chang, Tian Sheuan},
  journal={IEEE Transactions on Circuits and Systems I: Regular Papers},
  volume={65},
  number={5},
  pages={1642--1651},
  year={2017},
  publisher={IEEE}
}

@article{32,
  title={Angel-eye: A complete design flow for mapping CNN onto embedded FPGA},
  author={Guo, Kaiyuan and Sui, Lingzhi and Qiu, Jiantao and Yu, Jincheng and Wang, Junbin and Yao, Song and Han, Song and Wang, Yu and Yang, Huazhong},
  journal={IEEE transactions on computer-aided design of integrated circuits and systems},
  volume={37},
  number={1},
  pages={35--47},
  year={2017},
  publisher={IEEE}
}

@article{33,
  title={An efficient FPGA-based convolutional neural network for classification: Ad-MobileNet},
  author={Bouguezzi, Safa and Fredj, Hana Ben and Belabed, Tarek and Valderrama, Carlos and Faiedh, Hassene and Souani, Chokri},
  journal={Electronics},
  volume={10},
  number={18},
  pages={2272},
  year={2021},
  publisher={MDPI}
}

@article{35,
  title={A resource-limited hardware accelerator for convolutional neural networks in embedded vision applications},
  author={Moini, Shayan and Alizadeh, Bijan and Emad, Mohammad and Ebrahimpour, Reza},
  journal={IEEE Transactions on Circuits and Systems II: Express Briefs},
  volume={64},
  number={10},
  pages={1217--1221},
  year={2017},
  publisher={IEEE}
}

@article{n1,
author = {Alsharabi, Naif},
year = {2023},
month = {11},
pages = {12},
title = {Real-Time Object Detection Overview: Advancements, Challenges, and Applications},
volume = {3},
journal = {مجلة جامعة عمران},
doi = {10.59145/jaust.v3i6.73}
}

@article{n2,
author = {Chen, Xi and Guhl, Jan},
year = {2018},
month = {01},
pages = {149-154},
title = {Industrial Robot Control with Object Recognition based on Deep Learning},
volume = {76},
journal = {Procedia CIRP},
doi = {10.1016/j.procir.2018.01.021}
}

@article{n3,
title = {A review on edge analytics: Issues, challenges, opportunities, promises, future directions, and applications},
journal = {Digital Communications and Networks},
year = {2022},
issn = {2352-8648},
doi = {https://doi.org/10.1016/j.dcan.2022.10.016},
url = {https://www.sciencedirect.com/science/article/pii/S2352864822002255},
author = {Sabuzima Nayak and Ripon Patgiri and Lilapati Waikhom and Arif Ahmed}
}

@article{n4,
author = {Hasan, Mohammad Kamrul and Jahan, Nusrat and Ahmad Nazri, Mohd Zakree and Islam, Shayla and Khan, Muhammad and Alzahrani, Ahmed and Alalwan, Nasser and Nam, Yunyoung},
year = {2024},
month = {02},
pages = {1-1},
title = {Federated Learning for Computational Offloading and Resource Management of Vehicular Edge Computing in 6G-V2X Network},
volume = {PP},
journal = {IEEE Transactions on Consumer Electronics},
doi = {10.1109/TCE.2024.3357530}
}

@misc{link2,
    author = {{Hailo}},
    title = {{Hailo AI: The World’s Best Edge AI Processors}},
    howpublished = {\url{https://hailo.ai/products/ai-accelerators/hailo-8-ai-accelerator/}},
    note = {Online; accessed 1 July 2024}}

@unknown{youvan2024,
author = {Youvan, Douglas},
year = {2024},
month = {06},
pages = {},
title = {Developing and Deploying AI Applications on NVIDIA Jetson Orin NX: A Comprehensive Guide},
doi = {10.13140/RG.2.2.15641.43363}
}

@inproceedings{Nv2,
  title={Benchmarking Jetson Edge Devices with an End-to-end Video-based Anomaly Detection System},
  author={Pham, Hoang V and Tran, Thinh G and Le, Chuong D and Le, An D and Vo, Hien B},
  booktitle={Future of Information and Communication Conference},
  pages={358--374},
  year={2024},
  organization={Springer}
}
\vskip -1\baselineskip plus -1fil
\begin{IEEEbiography}
[{\includegraphics[width=1.2in,height=1.3in,clip,keepaspectratio]{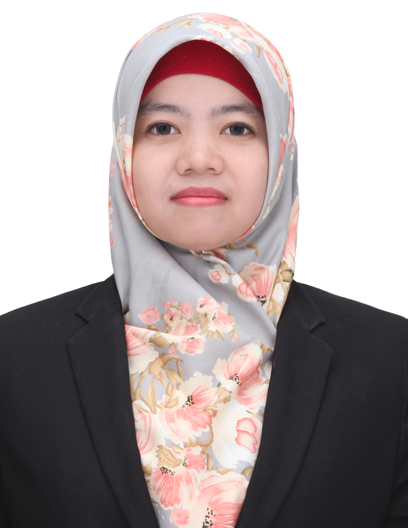}}]{Mas Nurul Achmadiah}{\space}received the Bachelor of Applied Science degree in Mechatronics Engineering from the Electronic Engineering Polytechnic Institute of Surabaya, Surabaya, Indonesia, in 2014. The Master's degree in Electronic Engineering from the Tenth of November Institute of Technology, Surabaya, Indonesia, in 2017, and is currently studying for a Ph.D. degree in the Electro-Optics Department at the National Formosa University in Yunlin, Taiwan. Also, a lecturer in the Department of Electrical Engineering, State Polytechnic of Malang, Indonesia. Her research interests are in artificial intelligence, Edge Computing Devices, intelligent control, and image processing.
\end{IEEEbiography}
\vskip -2\baselineskip plus -1fil
\begin{IEEEbiography}[{\includegraphics[width=1.2in,height=1.3in,clip,keepaspectratio]{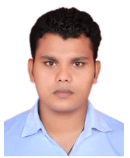}}]{Afaroj Ahamad}{\space} received his Ph.D. in Deep Learning Algorithms for Edge AI from National Formosa University, Huwei, Taiwan. He also holds an M.Tech. in Electronics Engineering from Vel Tech Rangarajan Dr. Sagunthala R\&D Institute of Science and Technology, Chennai, India, and a B.Tech. in Electronics and Communication Engineering from Aligarh Muslim University, Aligarh, India. He is currently an Assistant Professor in the Department of Computer Science and Engineering at Yuan Ze University, Taoyuan City, Taiwan. Previously, Dr. Ahamad served as an Assistant Professor in the Department of Computer Science and Information Engineering at Asia University, Taichung, Taiwan, from October 2024 to January 2025. From January 2023 to September 2024, he worked as a Senior FPGA Application Engineer at E-Elements Technology Co. Ltd., where he contributed to embedded system design and AI acceleration using DPU and FPGA technologies. His research interests include deep learning, machine learning (ML), FPGA-based AI acceleration, and IoT systems for edge computing. He has extensive experience in both academic and industrial settings, with a focus on deploying efficient AI solutions for IoT and robotics applications.

\end{IEEEbiography}
 \vskip -2\baselineskip plus -1fil
\begin{IEEEbiography}[{\includegraphics[width=1.2in,height=1.3in,clip,keepaspectratio]{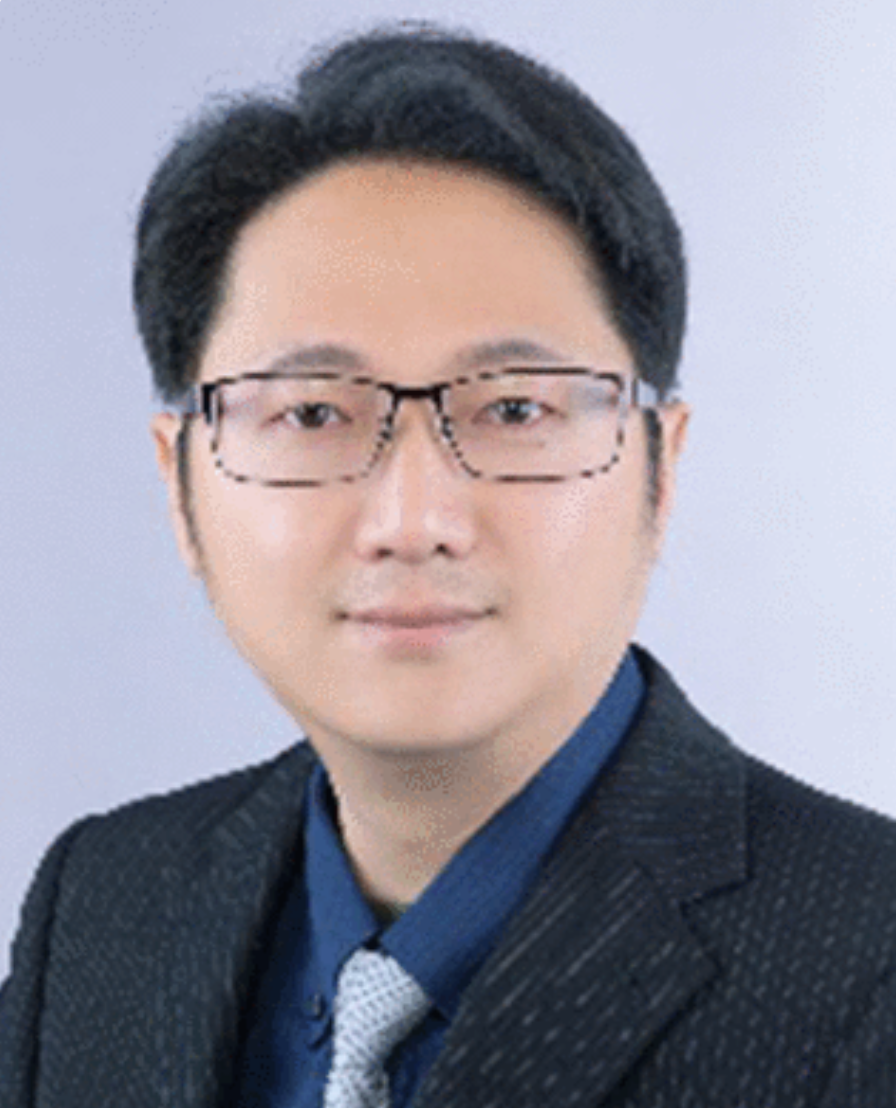}}]{Chi-Chia Sun}{\space}(Member, IEEE) received the B.S. degree in computer science and engineering from the National Taiwan Ocean University, Keelung City, Taiwan, in 2004, the M.S. degree in electronic engineering from the National Taiwan University of Science and Technology, Taipei, Taiwan, in 2006, and the Doktor Ingenieur degree from Dortmund University of Technology, Dortmund, Germany, in 2011. From April 2008 to March 2011, he worked as a Research Assistant with Dortmund University of Technology. Currently, he is a Full Professor with the Department of Electrical Engineering, National Taipei University, Taipei, Taiwan. His research interests include image processing, system integration, and very large scale integration (VLSI)/FPGA design.
\end{IEEEbiography}
\vskip -2\baselineskip plus -1fil

\begin{IEEEbiography}[{\includegraphics[width=1.2in,height=1.3in,clip,keepaspectratio]{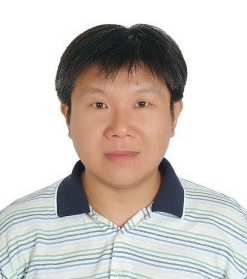}}]{Wen-Kai Kuo}{\space}(Member, IEEE) received a PhD in electronic engineering from the National Chiao Tung University, Hsin-Chu, Taiwan, in 2000. He has been a professor at the Department of Electro-optics Engineering at National Formosa University, Huwei, Yunlin, Taiwan. He is a member of the Phi-Tau-Phi Honorary Scholar Society. His research interests are optical sensors and systems.
\end{IEEEbiography}

\end{document}